\documentclass[10pt,twocolumn,letterpaper]{article}

\usepackage[pagenumbers]{cvpr}              %

\usepackage[dvipsnames]{xcolor}
\usepackage[utf8]{inputenc}
\usepackage{times}
\usepackage{epsfig}
\usepackage{graphicx}
\usepackage{amsmath,mathtools}
\usepackage{amssymb}
\usepackage{amsthm}
\usepackage{nicefrac}       %
\usepackage{microtype}      %
\usepackage{amsmath} 
\usepackage{float}
\usepackage{amsfonts}
\usepackage{bm}
\usepackage{enumitem}
\usepackage{mathtools}
\usepackage{multirow}
\usepackage{siunitx}
\usepackage[T1]{fontenc}

\usepackage{fontawesome}
\usepackage{pifont}
\usepackage{color}
\usepackage[algo2e,inoutnumbered,linesnumbered,algoruled,vlined]{algorithm2e}
\usepackage{algpseudocode}
\usepackage{booktabs}
\usepackage{tabularx}
\usepackage{bm}
\usepackage{url}
\usepackage{bbm}
\usepackage{threeparttable}
\usepackage{tikz}
\usepackage{wrapfig}
\usetikzlibrary{backgrounds}
\usetikzlibrary{decorations.pathreplacing}
\usepackage[outline]{contour}
\usepackage{appendix}
\definecolor{cvprblue}{rgb}{0.21,0.49,0.74}
\usepackage[pagebackref,breaklinks,colorlinks,citecolor=cvprblue]{hyperref}

\usepackage[accsupp]{axessibility}  %

\newcommand{\ourModel}{\textsc{Gen3C}\xspace}

\DeclareMathOperator*{\argmin}{argmin}

\newcommand{\parahead}[1]{\vspace{1mm}\noindent\textbf{#1.}\ }

\theoremstyle{definition}

\DeclarePairedDelimiterX{\infdivx}[2]{(}{)}{%
  #1\;\delimsize\|\;#2%
}

\definecolor{darkblue}{RGB}{49,130,189}
\definecolor{stanfordgrey}{RGB}{46,45,41}
\definecolor{cardinalred}{RGB}{253,141,60}

\usepackage[capitalize]{cleveref}

\crefname{section}{\S}{\S\S}
\crefname{subsection}{\S}{\S\S}
\crefname{conj}{Conj.}{Conj.}

\Crefname{assumption}{\textbf{H}\hspace{-3pt}}{\textbf{H}\hspace{-3pt}}
\crefname{assumption}{\textbf{H}}{\textbf{H}}

\crefname{algorithm}{\text{Alg.}}{\text{Alg.}}
\crefname{assumption}{\textbf{H}}{\textbf{H}}
\crefname{equation}{\text{Eq}}{\text{Eq}}
\crefname{definition}{\text{Dfn.}}{\text{Dfn.}}
\crefname{lemma}{\text{Lemma}}{\text{Lemma}}
\crefname{dfn}{\text{Dfn.}}{\text{Dfn.}}
\crefname{thm}{\text{Thm.}}{\text{Thm.}}
\crefname{tab}{\text{Tab.}}{\text{Tab.}}
\crefname{fig}{\text{Fig.}}{\text{Fig.}}
\crefname{table}{\text{Tab.}}{\text{Tab.}}
\crefname{figure}{\text{Fig.}}{\text{Fig.}}

\definecolor{mygreen}{RGB}{159, 200, 59}
\definecolor{myred}{RGB}{223, 135, 102}

\newcommand{\Point}{\mathbf{P}}
\newcommand{\Time}{t}
\newcommand{\View}{v}
\newcommand{\nView}{V}
\newcommand{\RGBImage}{I}
\newcommand{\Mask}{M}
\newcommand{\RGBVideo}{\mathbf{I}}
\newcommand{\MaskVideo}{\mathbf{M}}

\newcommand{\Render}{\mathcal{R}}
\newcommand{\Camera}{\mathbf{C}}
\newcommand{\videoLength}{L}

\newcommand{\depthEstimation}{\mathbf{d}}
\newcommand{\depthScale}{s}
\newcommand{\depthTranslation}{t}

\newcommand{\videoInput}{\mathbf{x}}
\newcommand{\diffusionNoise}{\mathbb{\epsilon}}
\newcommand{\diffusionTime}{\tau}
\newcommand{\diffusionTarget}{\mathbf{y}}

\newcommand{\diffusionCond}{\mathbf{c}}
\newcommand{\diffusionModel}{\mathbf{f}}
\newcommand{\diffusionModelParams}{\theta}
\newcommand{\videoOutput}{\hat{\mathbf{x}}}
\newcommand{\latent}{\mathbf{z}}
\newcommand{\vaeEncoder}{\mathcal{E}}
\newcommand{\vaeDecoder}{\mathcal{D}}
\newcommand{\dataDistribution}{p_{\text{data}}}

\usepackage[dvipsnames]{xcolor}

\definecolor{myblue}{RGB}{184, 218, 253}
\definecolor{myyellow}{RGB}{222,215,170}
\definecolor{mypink}{RGB}{250, 191, 213}

\def\paperID{3382} %
\def\confName{CVPR}
\def\confYear{2025}

\title{\ourModel : 3D-Informed World-Consistent Video Generation \\ with Precise Camera Control}

\author{Xuanchi Ren\footnotemark[1]~~$^{1,2,3}$ \;\; Tianchang Shen\footnotemark[1]~~$^{1,2,3}$ \;\; Jiahui Huang$^1$ \;\; Huan Ling$^{1,2,3}$ \;\; Yifan Lu$^1$ \;\; 
\\ Merlin Nimier-David$^1$ \;\; Thomas Müller$^1$ \;\; Alexander Keller$^1$ \;\; Sanja Fidler$^{1,2,3}$ \;\; Jun Gao$^{1,2,3}$ \\
 $^1$NVIDIA \quad $^2$University of Toronto \quad $^3$Vector Institute
}

\begin{document}
\twocolumn[{%
\renewcommand\twocolumn[1][]{#1}%
\maketitle
\begin{center}
\vspace{-9mm}
\renewcommand\arraystretch{0.5} 
\centering
\includegraphics[width=\linewidth]{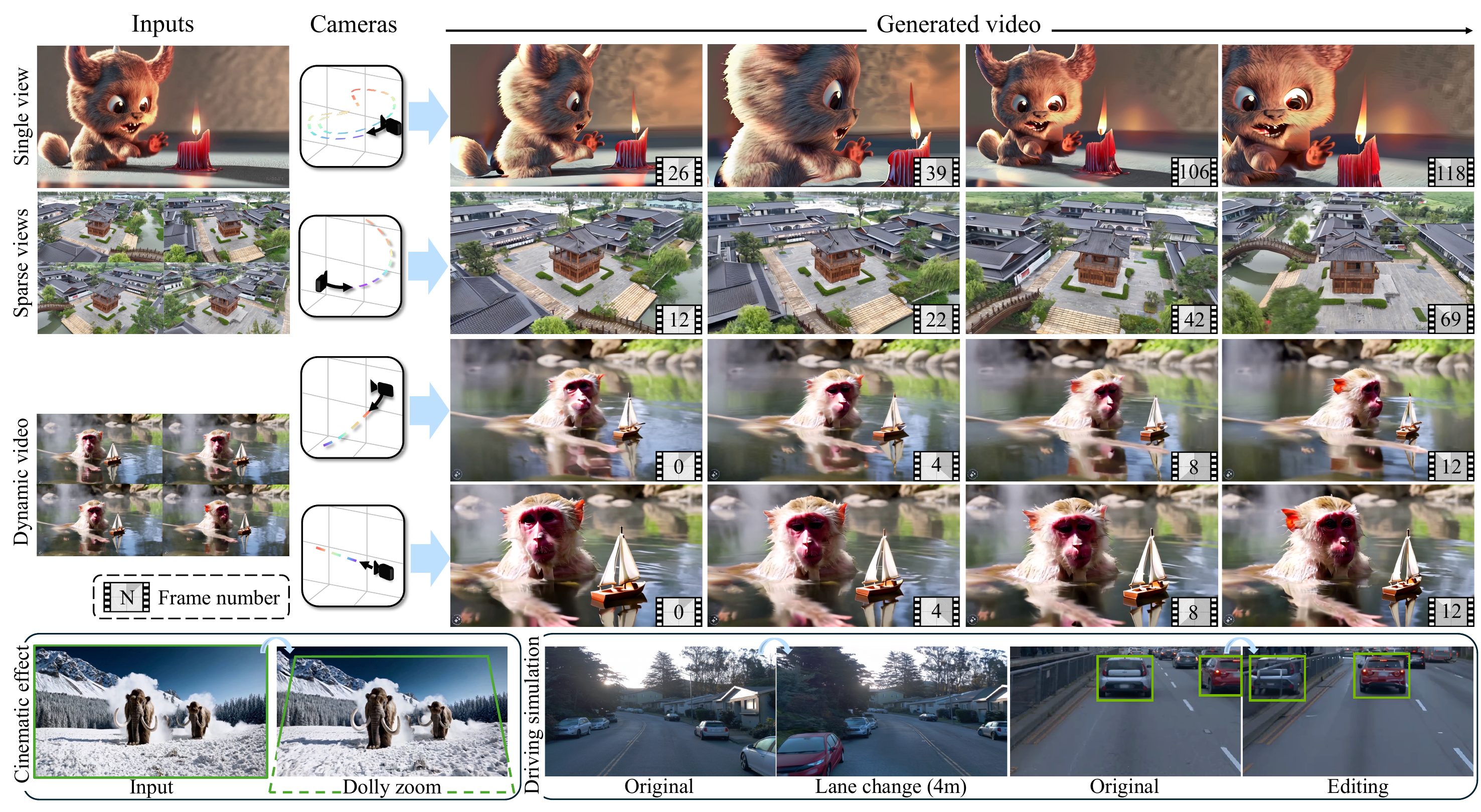}
\vspace{-6mm}
\captionof{figure}{\footnotesize
\ourModel can generate long and temporally consistent videos with precise camera control. We apply it to various applications, including single-view and sparse-views novel view synthesis, monocular dynamic video novel view synthesis, and driving simulation. With an explicit 3D cache, \ourModel further supports generating videos with cinematic effects, such as Dolly Zoom which 
 simultaneously changes poses and intrinsics, and 3D editing.
}
\vspace{-2mm}
\label{fig:teaser}
\end{center}
}]

\begin{abstract}
We present \ourModel, a generative video model with precise \textbf{C}amera \textbf{C}ontrol and temporal 3D \textbf{C}onsistency.
Prior video models already generate realistic videos, but they tend to leverage little 3D information, 
leading to inconsistencies, such as objects popping in and out of existence.
Camera control, if implemented at all, is imprecise, because camera parameters are mere inputs to the neural network which must then infer how the video depends on the camera.
In contrast, \ourModel is guided by a 3D cache: point clouds obtained by predicting the pixel-wise depth of seed images or previously generated frames.
When generating the next frames, \ourModel is conditioned on the 2D renderings of the 3D cache with the new camera trajectory provided by the user.
Crucially, this means that \ourModel neither has to remember what it previously generated nor does it have to infer the image structure from the camera pose.
The model, instead, can focus all its generative power on previously unobserved regions, as well as advancing the scene state to the next frame.
Our results demonstrate more precise camera control than prior work, as well as state-of-the-art results in sparse-view novel view synthesis, even in challenging settings such as driving scenes and monocular dynamic video.
Results are best viewed in videos. Check out our webpage! \url{https://research.nvidia.com/labs/toronto-ai/GEN3C/}

\end{abstract}

\vspace{-6mm}
\section{Introduction}
\label{sec:introduction}
\vspace{-1mm}

Creating immersive visual renderings that convey real-world scenery while enabling flexible viewing, manipulation and simulation thereof, is a longstanding aspiration in computer graphics, supporting industries including movie production, VR/AR, robotics and social platforms. 
However, traditional graphics workflows entail extensive manual effort and time in asset creation and scene design.
Recently, Novel View Synthesis (NVS) methods~\cite{mildenhall2021nerf,kerbl3Dgaussians} unleash this requirement and successfully produce realistic images at novel viewpoints of a scene with a set of posed images. However, such methods generally require dense input images and often suffer from severe artifacts when viewing from extreme viewpoints. 

\begin{figure}
  \centering
  \includegraphics[width=\linewidth]{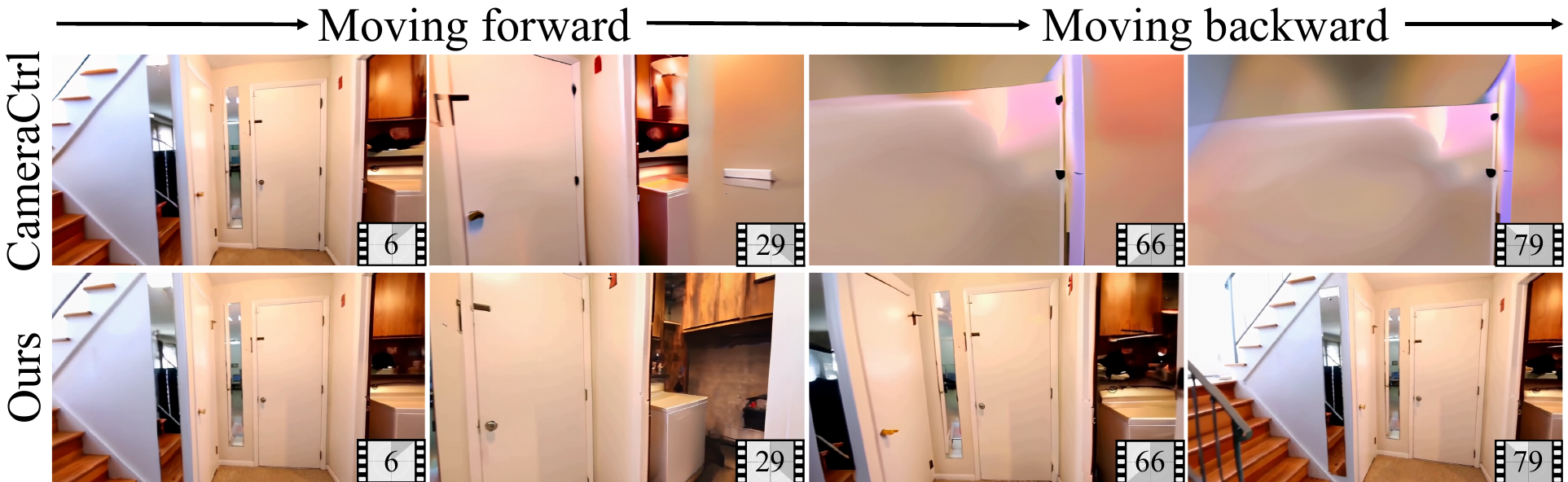}
  \vspace{-6mm}
  \caption{\footnotesize \textbf{Motivation:} Our model can generate consistent videos when the camera covers the same region multiple times, while previous work produces severe artifacts due to the lack of explicit modeling of the history.}
  \label{fig:motivation}
    \vspace{-5mm}
\end{figure}

More recently, video generation models, which can ``render'' photorealistic videos from text prompts, have demonstrated impressive visual quality and powerful content creation capabilities~\cite{sora, moviegen, runway,blattmann2023stable}, capturing the underlying distribution of real-world videos by training with massive amounts of data.
However, the key challenge towards practical applications in digital content creation workflows 
is controllability and consistency, \ie allowing the user to adjust camera motion, scene composition and dynamics, and maintaining spatial and temporal consistency across long-generated videos.
While several methods have been proposed to address this challenge through fine-tuning with images, additional text prompts or camera parameters~\cite{CameraCtrl,MotionCtrl,liu2023zero1to3,vanhoorick2024gcd,shi2023zero123plus,watson2024controlling}, 
achieving precise control for subtle or complex camera movements or scene arrangement remains unsolved. The model can easily forget previously generated content when looking back and forth; see Fig.~\ref{fig:motivation}.

Controllability and consistency in graphics pipelines are fundamentally rooted in their explicit modeling of 3D geometry and rendering it into 2D views. 
In this work, we take an initial step towards building this insight into the video generation models,
and propose \ourModel, a world-\textbf{C}onsistent video generation model with precise \textbf{C}amera \textbf{C}ontrol.
Its core is an approximated 3D geometry—akin to 3D modeling in graphics pipelines—constructed from user-provided images,
and can be precisely projected to any camera trajectory to guide video generation, providing strong conditioning for visual consistency.
In addition, ``rendering'' with video generation models leverages the rich prior from pre-trained large models, enabling NVS in sparse-view settings.

Specifically, we construct a 3D cache, represented as a point cloud, by unprojecting a depth estimate of the input image(s) or previously generated video frames. 
With the camera trajectory from the user, we then render the 3D cache and use the rendered video as conditioning input for the video model.
The video model is fine-tuned to translate imperfectly rendered video into a high-quality video, correcting any artifacts that stem from the 3D unprojection-projection process and filling in missing information.
This way, we achieve precise control of the camera and encourage the generated video to remain consistent over time. 
When multiple views are provided, we maintain a separate 3D cache for each individual view and leverage the video model to handle potential misalignment 
and aggregattion across views. 
Acting as an explicit geometry, the 3D cache further enables scene manipulation by simply modifying the 3D point cloud. 
  
We extensively evaluate our model on video generation tasks with varying input conditions, ranging from a single image to sparse and dense multi-view inputs. Our model generalizes well to dynamic scenes and demonstrates the ability to accurately control viewpoint, generate 3D-consistent high-fidelity videos, and fill in occluded or missing regions in the 3D cache. Beyond novel view synthesis, we explore applications enabled by the explicit 3D cache, such as object removal, and scene editing. 
We believe these results validate our approach as a step toward applying video generation models in production and simulation environments.

\vspace{-1mm}
\section{Related Work}
\label{sec:related}
\vspace{-1mm}

\parahead{Novel View Synthesis (NVS)}
Generating novel views from a set of posed images has seen significant progress~\cite{nerfstudio,mildenhall2021nerf,kerbl3Dgaussians}, with numerous extensions towards large scene reconstruction~\cite{barron2022mipnerf360,yang2023emernerf,Yu2024GOF,Li_2023_CVPR}, improved rendering quality~\cite{adaptiveshells2023,Huang2DGS2024,barron2021mipnerf}, faster rendering speed~\cite{adaptiveshells2023,mueller2022instant} and handling dynamic scenes~\cite{luiten2023dynamic,duan:2024:4drotorgs}. 
Yet, many of these methods require a dense set of input images and may produce severe artifacts when viewed from extreme viewpoints.
Several works proposed to address these issues through regularization using geometric priors~\cite{Yu2022MonoSDF,Niemeyer2021Regnerf,Yang2023FreeNeRF,somraj2023simplenerf,deng2022depth, roessle2022dense,wang2023sparsenerf,zhu2023FSGS}, which, however, are sensitive to noise in the estimated depth or normals.
Alternative approaches seek to train a feed-forward model to predict novel views from sparse posed images~\cite{yu2020pixelnerf,mvsnerf,zhou2023nerflix,tang2024lgm,jin2024lvsmlargeviewsynthesis,wang2021ibrnet,pixelsplat,chen2025mvsplat,ren2024scube}, but these methods are limited by the scarcity of training data and struggle to generalize to unseen domains and extreme novel views. 
With the recent success of image/video generation models,
ReconFusion~\cite{wu2024reconfusion} and CAT3D~\cite{gao2024cat3d} started leveraging prior knowledge learned by these models to facilitate sparse-view NVS. Due to the necessity of per-scene optimization, these methods remain inherently slow. Concurrent and unpublished work, including ReconX~\cite{liu2024reconx} and ViewCrafter~\cite{yu2024viewcrafter}, are closer to our work.
However, they rely on the alignment of input multiple views using DUSt3R~\cite{dust3r_cvpr24}, which is not robust to thin structures,
and introduces artifacts when misalignment happens. 
MultiDiff~\cite{Muller_2024_CVPR} leverages depth to warp a single image as guidance for novel view synthesis using a video diffusion model. It, however, only focuses on single view setting.

\begin{figure*}
  \centering
  \vspace{-4mm}
  \includegraphics[width=\linewidth]{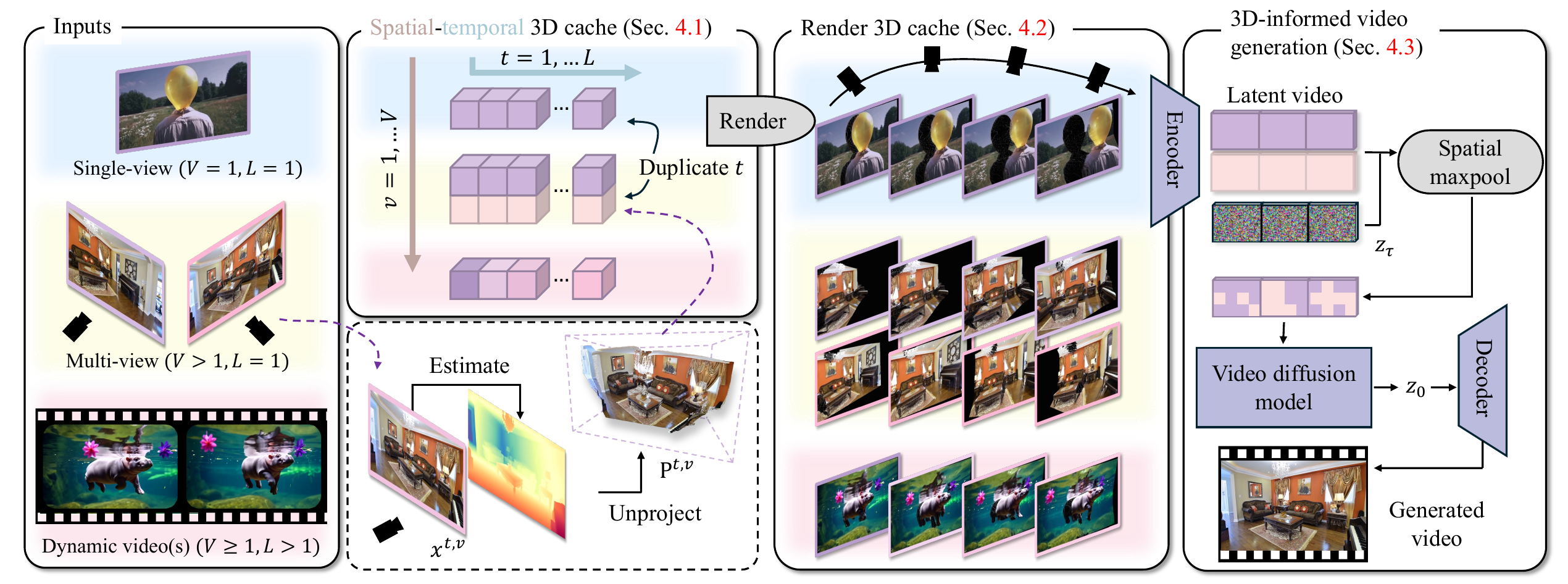}
  \vspace{-7mm}
  \caption{\footnotesize
    Overview of \ourModel. With the user input, which can be a \textcolor{myblue}{single-view} image, \textcolor{myyellow}{multi-view} images, or \textcolor{mypink}{dynamic video(s)}, we first build a spatiotemporal 3D cache (Sec.~\ref{sec:build_3d_cache}) by predicting the depth for each image and unprojecting it into 3D. With the camera poses from the user, we then render the cache into video(s) (Sec.~\ref{sec:rendering}), which are fed into the video diffusion model to generate a photorealistic video that aligns with the desired camera poses (Sec. \ref{sec:fuse_3D_cache} \& \ref{sec:model_training_method}). 
  }
  \vspace{-5mm}
  \label{Fig:Pipeline}
\end{figure*}

\parahead{Camera-Controllable Video Generation}
Early works propose inputting numerical camera parameters into their video generation models as an additional condition to fine-tune for camera control~\cite{liu2023zero1to3,shi2023zero123plus,vanhoorick2024gcd,guo2023animatediff,MotionCtrl,watson2024controlling}. 
However, these works struggle with precise control due to the model having to learn the mapping from the camera parameters to video, usually failing to generalize to camera motion that is different from the training data. 
Several training-free methods~\cite{NVS_Solver,camtrol} proposed to leverage depth to warp a single frame to a given camera trajectory and incorporate the result in the denoising process of a pre-trained diffusion model. This requires tuning the degree of consistency between the depth-warped images and the denoising output, leading to either artifacts or imprecise camera control.

\parahead{Consistent Video Generation}
Early work~\cite{mallya2020world} leverages a 3D point cloud, similar to our 3D cache, obtained by applying structure from motion to past frames. Renders of this point cloud are used to condition a Generative Adversarial Network~\cite{goodfellow2020generative}.
Instead, we estimate the depth for each seed image that is then reconciled by a diffusion-based video generation model, which is more robust to small-overlap images.
Streetscapes~\cite{deng2024streetscapes} improved the consistency of video diffusion models by relying on a precise height map of the environment that is not necessarily available. More recently, 
CVD~\cite{kuang2024collaborative} make synchronous frames of generated videos consistent with each other. However, overall consistency is still lost if content temporarily leaves the view of \emph{all} videos, because no history is maintained.
StreamingT2V~\cite{streamingt2v} maintain a history in the form of latent feature maps to enhance consistency, but camera control remains difficult because the history is latent rather than 3D.

\vspace{-1mm}
\section{Background: Video Diffusion Models}
\label{Sec:Background}
\vspace{-1mm}
As our method is based on a video diffusion model, we briefly review their principles.
A diffusion model $\diffusionModel_{\diffusionModelParams}$ learns to model a data distribution $\dataDistribution(\videoInput)$ via an iterative denoising process.
To train the model, noisy versions $\videoInput_\diffusionTime = \alpha_\diffusionTime \videoInput_0 + \sigma_\diffusionTime \diffusionNoise$ of a data sample $\videoInput_0 \sim \dataDistribution(\videoInput)$ are constructed by adding noise $\diffusionNoise$ sampled from a Gaussian distribution, $\mathcal{N}(\mathbf{0}, \mathbf{I})$, with the noise schedule parameterized by $\alpha_\diffusionTime$ and $\sigma_\diffusionTime$. The diffusion time $\diffusionTime$ is sampled from the distribution $p_\tau$.
Then, the parameters $\diffusionModelParams$ of the diffusion model $\diffusionModel_{\diffusionModelParams}$ are optimized to minimize the denoising score matching objective function:
\begin{equation} \label{Eqn:DSMObjective}
\mathbb{E}_{\videoInput_0\sim\dataDistribution(\videoInput), \diffusionTime\sim p_{\diffusionTime}, \diffusionNoise\sim\mathcal{N}(\mathbf{0}, \mathbf{I})} \left[\|\diffusionModel_{\diffusionModelParams}(\videoInput_\diffusionTime; \diffusionCond, \diffusionTime) - \diffusionTarget\|_2^2\right],
\end{equation}
where $\diffusionCond$ is optional conditions, and the target $\diffusionTarget$ can be $\diffusionNoise$, $\alpha_\diffusionTime\diffusionNoise - \sigma_\diffusionTime\videoInput_0$~\cite{FastSamplingDM}, or $\videoInput_0$~\cite{karras2022elucidating}, depending on the selected denoising process. Once trained, iteratively applying $\diffusionModel_{\diffusionModelParams}$ to a sample of Gaussian noise will produce a sample of $\dataDistribution(\videoInput)$.

In diffusion-based video generation models, latent diffusion models~\cite{blattmann2023align} are frequently employed to compress the video for operation in a lower-dimensional space.
Specifically, given the a RGB video data $\videoInput\in\mathbb{R}^{\videoLength\times 3 \times H \times W}$, where $\videoLength$ is the number of frames of size $H \times W$, a pre-trained VAE encoder $\vaeEncoder$ will encode the video into a latent space, i.e. $\latent = \vaeEncoder(\videoInput) \in\mathbb{R}^{\videoLength^\prime \times C \times h \times w}$.
Training and inference of the diffusion model are performed in this latent space. The final video $\videoOutput = \vaeDecoder(\latent)$ is decoded with a pre-trained VAE decoder $\vaeDecoder$.
In this paper, we leverage the pre-trained Stable Video Diffusion~\cite{blattmann2023stable} model, which is conditioned on an image $\diffusionCond$ and only compresses the video in the spatial dimension:
$\videoLength^{\prime} = \videoLength, C = 4, h = \frac{H}{8}$, and $w = \frac{W}{8}$.
However, our method is compatible with any other image-to-video diffusion model, as it does not rely on details of its architecture.

\vspace{-1mm}
\section{Method: 3D-Informed Video Generation}
\label{sec:method}
\vspace{-1mm}
Our key idea is to use 3D guidance to inform video generation, enabling precise camera control and improving consistency across the video frames. 
For this purpose, we first build a 3D cache from the input image(s) or pre-generated video frames (Sec.~\ref{sec:build_3d_cache}).
Then, the 3D cache will be rendered into the camera plane with the camera poses from the user (Sec.~\ref{sec:rendering}). 
Such renderings, while imperfect, provide a strong condition for the video generation model on the visual content it needs to generate (Sec.~\ref{sec:fuse_3D_cache}).
Our video generation model is fine-tuned accordingly to produce a 3D-consistent video that precisely aligns with the desired camera poses (Sec.~\ref{sec:model_training_method}). 
Fig.~\ref{Fig:Pipeline} provides an overview of our method.

\vspace{-1mm}
\subsection{Building a Spatiotemporal 3D Cache}
\label{sec:build_3d_cache}
\vspace{-1mm}
Choosing a proper 3D cache compatible with different applications and that generalizes to different scenes is the main consideration in our design. 
Recently, depth estimation has achieved significant progress across various domains~\cite{depth_anything_v2,hu2024depthcrafter,ke2023repurposing,depth_anything_v1,dust3r_cvpr24}, such as indoor, outdoor, or self-driving scenes. We thus choose the colored point cloud, unprojected from the depth estimation of an RGB image, as the basic element of our 3D cache. 

Specifically, we maintain a spatiotemporal 3D cache.
For an RGB image seen from a camera viewpoint $\View$ at time $\Time$, we create a point cloud, $\Point^{\Time,\View}$, by unprojecting the depth estimation of this RGB image.
We denote the number of camera views as $\nView$, and the length in temporal dimension as $\videoLength$; thus, our 3D cache is an $\videoLength \times \nView$ array of point clouds.

We build the spatiotemporal 3D cache according to the specific downstream applications. 
For {\em single image to video generation}, we only create one cache element ($\nView=1$) for the given image, and duplicate it $\videoLength$ times along the temporal dimension to generate the video of length $\videoLength$.
For {\em static NVS}, we create one cache element for each of the $\nView$ images provided by the user, and duplicate them $\videoLength$ times along the temporal dimension. We can then enable both sparse-view and dense-view NVS.
For {\em dynamic NVS}, we build the cache from each initial video(s) of identical length $L$ that are provided by a user or generated by another video model.  Then, $\nView$ equals to the number of time-synchronized videos,
and we can enable both single-view and multi-view dynamic NVS.
For these different applications, we assume the camera poses are provided along with the image(s) or video(s). If not, 
we estimate the camera poses using DROID-SLAM~\cite{teed2021droid}.

\begin{figure}
  \centering
  \includegraphics[width=\linewidth]{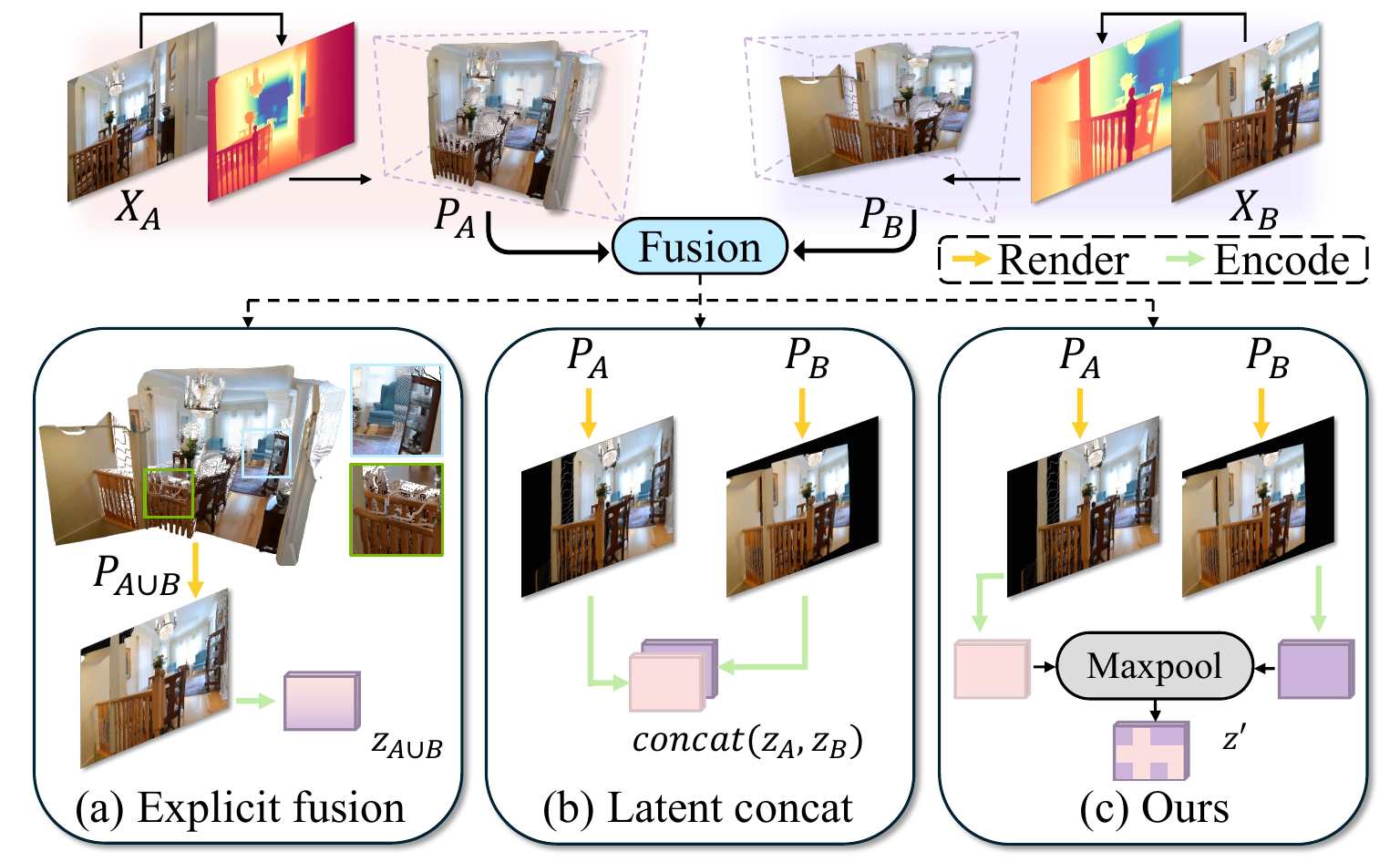}
  \vspace{-6mm}
  \caption{\footnotesize Three approaches to fuse the point cloud from two views.}
  \label{fig:injection_module_variants}
  \vspace{-5mm}
\end{figure}

Optionally, the 3D cache we build may be edited or simulated, for example, by removing or adding objects; we provide qualitative results in Sec.~\ref{sec:exp_dense_view}.
\subsection{Rendering the 3D Cache}
\label{sec:rendering}
Point clouds can be easily and efficiently rendered along any camera trajectory, much like Gaussian Splats~\cite{kerbl3Dgaussians}.
Such a rendering function $\Render$ maps $\Point^{\Time, \View}$ onto a tuple:
$
    (\RGBImage^{\Time, \View}, \Mask^{\Time, \View}) := \Render(\Point^{\Time, \View}, \Camera^\Time) ,
$
where $\RGBImage^{\Time, \View}$ is the RGB image as seen from a new camera $\Camera^\Time$.
The mask $\Mask^{\Time, \View}$ identifies disocclusions, flagging pixels that are not covered when rendering the point cloud. %
In that sense, the mask identifies regions of the image $\RGBImage^{\Time, \View}$ that need to be filled in.

For a sequence $\Camera=(\Camera^1, \ldots, \Camera^{\videoLength})$ of new camera poses from the user, we render all cache elements $\Point^{t, v}$ and obtain $\nView$ videos $(\Render(\Point^{1, \View}, \Camera^1), \Render(\Point^{2, \View}, \Camera^2),\ldots, \Render(\Point^{\videoLength, \View}, \Camera^\videoLength))$,
where $1 \leq \View \leq \nView$.
Concatenating the rendered images $( \RGBImage^{1, \View},\ldots, \RGBImage^{\videoLength, \View})$ and masks $( \Mask^{1, \View},\ldots, \Mask^{\videoLength, \View})$ along the temporal dimension for a camera view $\View$, we denote the resulting videos of the images and the masks by
$\RGBVideo^{\View}\in\mathbb{R}^{\videoLength\times 3\times H \times W}$ and $ \MaskVideo^{\View}\in\mathbb{R}^{\videoLength\times 1\times H \times W}$, respectively.

\begin{figure*}
  \centering
  \vspace{-7mm}
  \includegraphics[trim={0 0 0 0},clip,width=\linewidth]{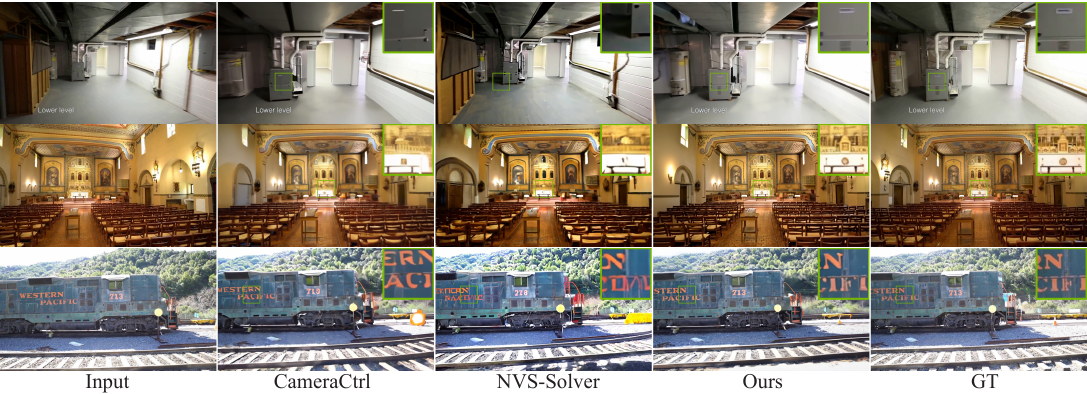}
  \vspace{-6mm}
  \caption{\footnotesize Qualitative results for single-view novel view synthesis. Compared to the baselines, our model generates photorealistic novel view images that precisely align with the camera poses. The green zoom-in boxes highlight the fine-grained details that our model can preserve.
  }
  \label{fig:single_view}
  \vspace{-6mm}
\end{figure*}

\subsection{Fusing and Injecting the 3D Cache}
\label{sec:fuse_3D_cache}
When conditioning a video diffusion model with the renderings of our 3D cache, the key challenge is that the 3D cache may be inconsistent across different camera viewpoints, either due to imperfect depth predictions or inconsistent lighting.
Hence, the model will need to aggregate the information (if $\nView > 1$) for a coherent prediction.
Our key principle when designing this module is to minimize the introduction of additional trainable parameters: as the pre-trained video diffusion model has been trained on massive Internet data, any new parameters may not generalize as well.

Specifically,
we modify the forward computation process 
of the image-to-video diffusion model, denoted by $\diffusionModel'_{\diffusionModelParams}$.
We first encode the rendered video $\RGBVideo^{\View}$ using the frozen VAE encoder $\vaeEncoder$ to obtain the latent video $\latent^{{\View}} = \vaeEncoder(\RGBVideo^{\View})$, 
and mask out the regions that are not covered by the 3D cache, as indicated by the masks $\MaskVideo^{\View}$.
During training, we
then concatenate the masked latent with the noisy version $\latent_\diffusionTime = \alpha_{\diffusionTime}\vaeEncoder(\videoInput) + \sigma_{\diffusionTime} \diffusionNoise$ of the target video $\videoInput$ in latent space along the channel dimension, and feed it into the video diffusion model. 
To fuse the information from multiple viewpoints, we separately feed each viewpoint into the first layer of the diffusion model, denoted by $\textit{In-Layer}$,
and apply max-pooling over all the viewpoints to get the final feature map. In summary:
\begin{align}
    \latent^{\View, \prime} &= \textit{In-Layer} (\text{Concat}(\latent^{\View} \odot  \MaskVideo^{\View,\prime}, \latent_\diffusionTime)), \label{eqn:concat_train}\\
    \latent^{\prime} &= \text{Max-Pool}\{\latent^{1, \prime}, \ldots, \latent^{\nView, \prime}\},
\end{align}
where $\odot$ denotes element-wise multiplication and $\MaskVideo^{\View,\prime}\in \mathbb{R}^{\videoLength\times 1 \times h \times w}$ is obtained by downsampling the $\MaskVideo^{\View}$ using min-pooling with size $\frac{H}{h} \times \frac{W}{w}$, in order to align with the latent dimension (Sec.~\ref{Sec:Background}). 
The resulting feature map $\latent^{\prime}$ is further processed in the video diffusion model
that is fine-tuned to generate a consistent video conditioned on these renderings from the 3D cache.

\parahead{Discussion}
The strategy described above is a general mechanism to aggregate the information from multiple views and inject it into the video diffusion model.
We compare it to alternatives that exhibit different properties, as illustrated in Fig.~\ref{fig:injection_module_variants}.
See Sec.~\ref{sec:ablation_study} for an empirical comparison.

The \textbf{explicit fusion} approach, proposed in concurrent works~\cite{liu2024reconx,yu2024viewcrafter}, directly fuses the point clouds in 3D space.
While being simple, such an approach strongly relies on depth alignment and will introduce artifacts when inconsistencies manifest across multiple viewpoints.
Furthermore, it would be nontrivial to imbue such fused cache with view-dependent lighting information.
For these reasons, we prefer to let the model handle aggregation of view information.
Another approach, which we denote as ``\textbf{concat}'', is to concatenate all latents for the rendered cache along the channel dimension. 
Although this approach empirically works well, it requires bounding the maximum number of viewpoints the model can support by a constant, and imposes an order on the viewpoints. 
Instead, we favored a permutation-invariant fusion operation, leading to our pooling-based strategy. %

Another key design choice is the incorporation of the masking information into the model. 
We initially tried to concatenate the mask channel to the latent. However, the concatenation operation introduces additional model parameters, which would now need to be trained, and therefore may not generalize well when the mask channel is not represented in any large-scale training data.
Instead, we apply the mask values directly to the latents by element-wise multiplication, leaving the model architecture unchanged. We provide emperical results of this observation in the Supplement.

\subsection{Model Training}
\label{sec:model_training_method}
With the renderings of the 3D cache as the conditioning signal $\diffusionCond$, we fine-tune our modified video diffusion model $\diffusionModel^\prime_{\diffusionModelParams}$.
Specifically, we first create pairs of the renderings of the 3D cache, $\Render(\Point^{\Time,\View}, \cdot)$, and the corresponding RGB ground truth video $\videoInput$ along a new camera trajectory from our training data.
We then fine-tune the video diffusion model using our fusion strategy (Sec.~\ref{sec:fuse_3D_cache}) with the denoising score matching objective function of Eqn.~(\ref{Eqn:DSMObjective}), where the target $\diffusionTarget$ is  $\latent_0=\vaeEncoder(\videoInput)$ following the pre-trained image-to-video diffusion model practices~\cite{blattmann2023stable}. We also encode the first frame using the CLIP model~\cite{radford2021learning} as an additional condition.
Details about creating paired data used for fine-tuning are provided in Sec.~\ref{sec:experiments}.

\begin{figure*}
  \centering
  \vspace{-9mm}
  \includegraphics[trim={0 0 0 150},clip,width=\linewidth]{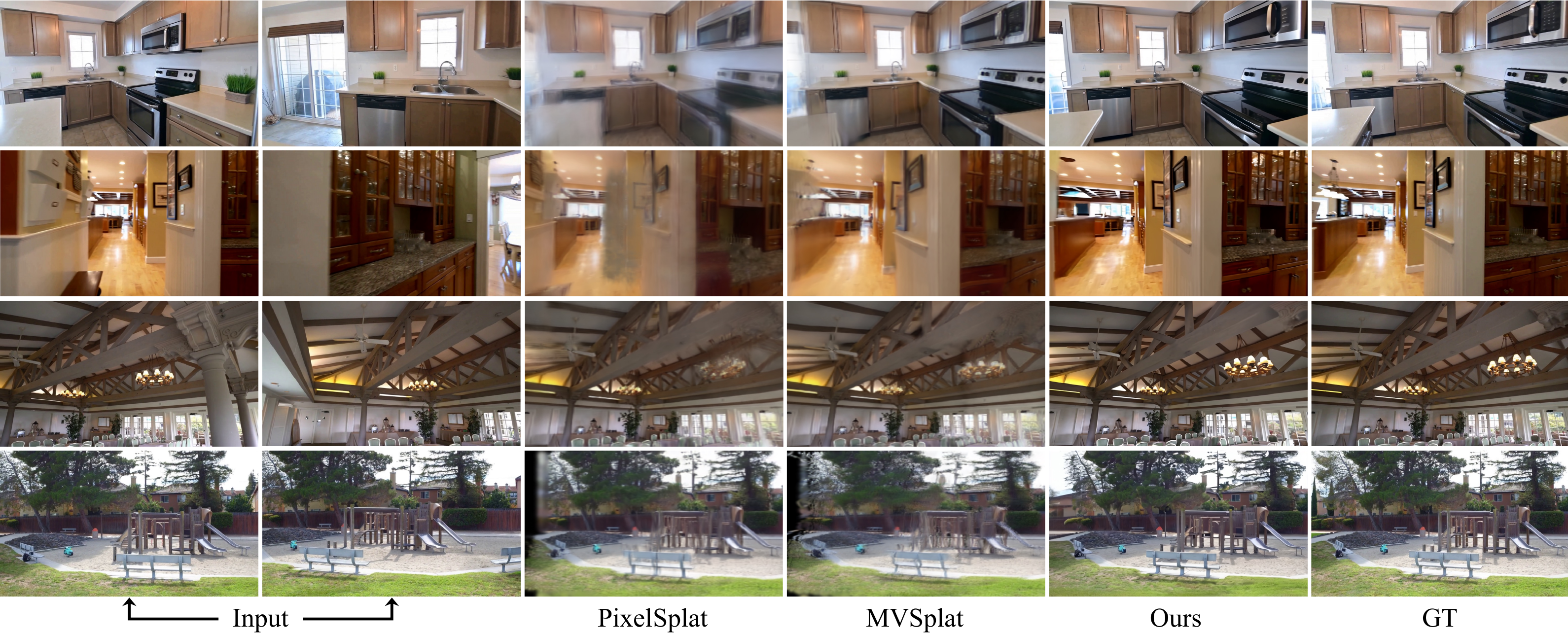}
  \vspace{-7mm}
  \caption{\footnotesize Qualitative results on two-views novel view synthesis. Compared to the baselines, our model generates much more plausible and realistic novel views with a smooth transition between two input views even if the overlap is small (such as the first row).
  }
  \label{fig:sparse_view}
  \vspace{-5mm}
\end{figure*}

\vspace{-1mm}
\subsection{Model Inference}
\label{sec:chunks}
\vspace{-1mm}
For inference, we initialize the latent code $\latent$ with Gaussian noise and iteratively denoise this latent code using our modified video diffusion model $\diffusionModel'_\diffusionModelParams$, conditioned on the renderings of our 3D cache. The final RGB video is obtained by running the pre-trained VAE decoder $\vaeDecoder$ on the denoised latent code. 
Generating a 14-frame video takes around 30 seconds on one A100 NVIDIA GPU. 

\parahead{Autoregressive Inference and 3D Cache Updates}
Many applications require generating long videos, but, the longer the video, existing models are particularly prone to inconsistencies.
To generate long, \emph{consistent} videos, we propose updating our 3D cache incrementally.
We first divide the long video into overlapping chunks of length $\videoLength$ with a one-frame overlap between two consecutive chunks.
We then render the 3D cache and generate the frames of each chunk in sequence autoregressively.
To make the prediction consistent over time, we update the 3D cache using previously generated chunks: for each generated frame in a chunk, we estimate its pixel-wise depth using a depth estimator~\cite{depth_anything_v2}.
Since the camera pose of the frames is known (user-provided), we can align the depth estimation with the existing 3D cache by minimizing reprojection error; details in the supplement.
The aligned RGB-D frame is then unprojected into 3D and concatenated with the 3D cache, which is subsequently used for predicting the next chunk of frames.

\vspace{-2mm}
\section{Experiments and Applications}
\label{sec:experiments}
\vspace{-1mm}
In this section, we introduce the experimental setup to train \ourModel (Sec.~\ref{sec:exp_setup})
and showcase its versatility with several downstream tasks, including single image to video generation (Sec.~\ref{sec:exp_single_view}), two-views NVS (Sec.~\ref{sec:exp_sparse_view}),  NVS for driving simulation (Sec.~\ref{sec:exp_dense_view}), and monocular dynamic NVS (Sec.~\ref{sec:exp_dynamic}).  We provide ablation studies in Sec.~\ref{sec:ablation_study}.

\subsection{Training Details}
\label{sec:exp_setup}
A key challenge in training \ourModel is the lack of multi-view, dynamic, real-world video data, that provides the pairs of 3D cache and ground truth video for a novel camera trajectory.
We leverage static real-world video to help the model reason about spatial consistency and synthetic multi-view dynamic video to help with temporal consistency. 

\parahead{Datasets} 
We select three real-world videos datasets: RE10K \cite{zhou2018stereo}, DL3DV \cite{ling2024dl3dv}, Waymo Open Dataset (WOD) \cite{sun2020scalability}, and a synthetic dataset Kubric4D \cite{vanhoorick2024gcd,greff2022kubric}. 
RE10K~\cite{zhou2018stereo} consists of 74,766 video clips that capture real-world real-estate scenes for both indoors and outdoors. 
We estimate camera parameters with DROID-SLAM~\cite{teed2021droid} and predict per-frame depth using DAV2~\cite{depth_anything_v2}. The depth prediction is aligned with the scene scale from DROID-SLAM.
DL3DV~\cite{ling2024dl3dv} contains 10k videos of real-world scenes.
We annotate these clips following the same protocol as RE10K.
WOD~\cite{sun2020scalability} is a real-world driving dataset with 1000 scenes and each scene has 200 frames. 
We use DAV2~\cite{depth_anything_v2} to predict the depth and rigidly align it with the scale from the LiDAR point cloud. 
For Kubric4D~\cite{greff2022kubric}, we use the 3000 scenes generated by GCD~\cite{vanhoorick2024gcd}, which includes multi-object dynamics. This dataset is in the format of point cloud sequences and we render RGB-D videos for desirable camera trajectories. 

\begin{table}[t]
\begin{center}
\resizebox{\linewidth}{!}{
\begin{tabular}{l@{\hspace{2mm}}c@{\hspace{2mm}}c@{\hspace{2mm}}c@{\hspace{2mm}}c@{\hspace{2mm}}c@{\hspace{2mm}}c@{\hspace{2mm}}c}
\toprule
& \multicolumn{3}{c}{\textbf{Tanks-and-Temples~\cite{knapitsch2017tanks}}} & \multicolumn{4}{c}{\textbf{RE10K~\cite{zhou2018stereo}}} \\
\cmidrule(lr){2-4} 
\cmidrule(lr){5-8} 
\multicolumn{1}{l}{Method} & PSNR $\uparrow$ & SSIM $\uparrow$ & LPIPS$\downarrow$ & PSNR $\uparrow$ & SSIM $\uparrow$ & LPIPS$\downarrow$  & TSED$\uparrow$\\
\midrule
MotionCtrl~\cite{MotionCtrl} & 13.46 & 0.46 & 0.42  & 13.60 & 0.59 & 0.46 &0.1363   \\
CameraCtrl~\cite{CameraCtrl} & 15.88 & 0.55 & 0.29 & 18.40 & 0.72 & 0.25 &0.8033  \\
GenWarp~\cite{GenWarp} & 16.04 & 0.50 & 0.39  & 15.50 & 0.61 & 0.40 & 0.0330 \\
NVS-Solver~\cite{NVS_Solver} & 16.95 & 0.59 & 0.27 & 16.90 & 0.69 & 0.30  & 0.7286\\
\ourModel & \textbf{18.66} & \textbf{0.67} & \textbf{0.20} & \textbf{19.88} & \textbf{0.78} & \textbf{0.20} & \textbf{0.9143}\\
\bottomrule
\end{tabular}
}
\end{center}
\vspace{-6.5mm}
\caption{\footnotesize Quantitative results for single view to video generation. \textit{RE10K} is the in-domain dataset and \textit{Tanks and Temples} is the out-of-domain dataset. 
}
\label{table:single}
\vspace{-7mm}
\end{table}

\parahead{Paired Data Curation}
For real-world videos, we train our model to predict current frames using past or future frames from the same sequence. In this way, we effectively simulate viewpoint changes, allowing the model to extrapolate to unseen viewpoints and generate consistent video informed by the observations. %
Specifically, 
for RE10K and DL3DV, we randomly select equally-spaced $\nView\in [1, 4]$ frames from the video clip to create our 3D cache. 
The ground truth video consists of $\videoLength$ consecutive frames that include one of the selected $\nView$ frames. 
For the WOD, we randomly sample a sequence of time-synchronized $\videoLength$ frames for each of the three cameras to create our 3D cache ($\nView=3$ in this case). The ground truth video is only sampled from the FRONT camera.
We use all three cameras to create the 3D cache because it allows the model to learn to resolve inconsistencies across cameras, such as depth prediction, camera ISP, etc. %
For Kubric, 
we render each dynamic scene from $\nView\in [1,4]$ camera trajectories to generate multi-view dynamic videos for creating the 3D cache. 
We then render one additional video at a different camera trajectory as ground truth. %
Although we choose maximum $\nView$ to be $4$ for training, it is flexible and can be an arbitrary number.

\begin{figure*}
  \centering
  \vspace{-10mm}
  \includegraphics[width=\linewidth]{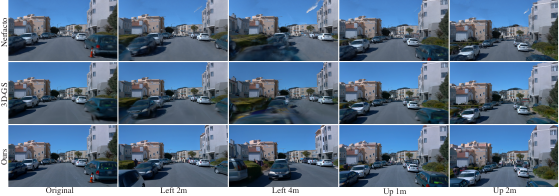}
  \vspace{-7mm}
  \caption{\footnotesize
  Qualitative results on novel view synthesis for driving scene. Our model can fill in the missing regions in the original video even when the deviation is large, while reconstruction-based baselines produce severe artifacts. 
  } 
    \label{fig:dense_view}
    \vspace{-5.5mm}
\end{figure*}

\subsection{Single View to Video Generation}
\label{sec:exp_single_view}
\ourModel can be easily applied to video/scene creation from a single image. We first predict the depth of the given image, then create the 3D cache, and render it into a 2D video, which is fed into the trained video diffusion model to generate a video that precisely follows the given camera trajectory.

\parahead{Evaluation and Baselines}%
We compare \ourModel to four baselines, including  GenWarp~\cite{GenWarp}, MotionCtrl~\cite{MotionCtrl},  CameraCtrl~\cite{CameraCtrl}, and NVS-Solver~\cite{NVS_Solver}. For a fair comparison to GenWarp~\cite{GenWarp} and NVS-Solver~\cite{NVS_Solver}, we use the same depth estimator~\cite{depth_anything_v2} to get the pixel-wise depth and rigidly align it by globally shifting and scaling using the scene scale. 
CameraCtrl~\cite{CameraCtrl} is the most related work, and we reproduce it by training with the same datasets, training protocol, and video diffusion model as in our method and replacing the rendered videos from our 3D cache with Pl\"ucker embeddings of camera trajectories.%
We evaluate all the methods on two datasets: \textit{RE10K}, which serves for in-domain testing.
and \textit{Tanks and Temples (T-\&-T)}, which serves for out-of-domain testing to evaluate generalization capabilities. To ensure a comprehensive evaluation, we sample 100 testing sequences for both \textit{RE10K} and \textit{T-\&-T}.
Following prior work~\cite{chen2025mvsplat,yu2024viewcrafter,DBLP:conf/cvpr/RenW22}, we report both pixel-align metrics, \ie, PSNR and SSIM~\cite{wang2004image}, and perceptual metrics, \ie, LPIPS~\cite{zhang2018unreasonable}. We further report TSED score~\cite{tsed} to evaluate the 3D consistency of the prediction.

\parahead{Results}
Quantitative results are provided in Table~\ref{table:single}.  Our method outperforms all the baselines in both out-of-domain and in-domain testing, demonstrating the strong capability of generating photorealistic videos from a single image.
Notably, Pl\"ucker-embedding based methods, such as Camera\-Ctrl~\cite{CameraCtrl}, generalize poorly to out-of-domain data, which has both different scene layouts and camera trajectories.
Thanks to the explicit modeling of 3D content in our 3D cache, our model only suffers a small performance drop.
We provide a qualitative comparison with the two strongest baselines in Fig.~\ref{fig:single_view} and with the others in the Supplement. The prediction from our method precisely follows the ground truth camera trajectory and captures fine-grained detail such as the chair legs or letter words. In particular,  CameraCtrl~\cite{CameraCtrl} fails to precisely follow the camera motion, as reasoning the scene layout only from the Pl\"ucker embedding is hard.

\subsection{Two-Views  Novel View Synthesis}
\label{sec:exp_sparse_view}

\begin{table}[t]
\begin{center}
\vspace{0mm}
\resizebox{0.8\linewidth}{!}{
\begin{tabular}{clccc}
\toprule
&\multicolumn{1}{l}{Method} & PSNR $\uparrow$ & SSIM $\uparrow$ & LPIPS$\downarrow$  \\
\midrule
\multirow{3}{*}{\rotatebox[origin=c]{90}{T-\&-T}}&PixelSplat~\cite{pixelsplat} & 21.34 / 17.45 & 0.70 / 0.65 & 0.42 / 0.46 \\
&MVSplat~\cite{chen2025mvsplat} & 20.90 / 16.08 & 0.70 / 0.63 & 0.39 / 0.44\\
& \ourModel & \textbf{22.22} / \textbf{20.51} & \textbf{0.76} / \textbf{0.72} & \textbf{0.14} / \textbf{0.16} \\
\midrule
\multirow{3}{*}{\rotatebox[origin=c]{90}{RE10K}}&PixelSplat~\cite{pixelsplat} & 19.74 / 16.95 & 0.75 / 0.70 & 0.40 / 0.44 \\
&MVSplat~\cite{chen2025mvsplat} & 21.40 / 15.51 & 0.78 / 0.69 & 0.30 / 0.37\\
&\ourModel &  \textbf{24.08} / \textbf{21.56} & \textbf{0.86} / \textbf{0.83} & \textbf{0.11} / \textbf{0.15} \\

\bottomrule
\end{tabular}
}
\end{center}
\vspace{-5mm}
\caption{\footnotesize
Quantitative results for two-views NVS. The two values in each table cell represent the interpolation and extrapolation results, respectively. 
}
\label{table:two_view}
\vspace{-4mm}
\end{table}

We further apply \ourModel to a challenging sparse-view novel view synthesis setting, where only two views are provided and we generate novel views from these two views.
Similar to Sec.~\ref{sec:exp_single_view}, we first predict the depth for each view, create the 3D cache, and use the camera trajectory to render it into two videos, which are fed into and fused by \ourModel to generate the output video.
Note that during inference, our model is not limited to two views and can be applied to any number of views. We provide qualitative results in the Supplement. 

\parahead{Evaluation and Baselines}
We compare our method to two representative works for sparse view reconstruction,  PixelSplat~\cite{pixelsplat} and MVSplat~\cite{chen2025mvsplat}. 
In this task, we evaluate both the interpolation and extrapolation capabilities of our model. Specifically, we randomly select two input frames from a video. For interpolation, we select target views between the input frames, and for extrapolation, we choose target views outside the range of the two input frames. 
We sample 40 testing sequences from both \textit{RE10K}~\cite{zhou2018stereo} and  \textit{T-\&-T}~\cite{knapitsch2017tanks}, and report PSNR, SSIM and LPIPS. 

\parahead{Results}
We provide quantitative results in Table~\ref{table:two_view} with qualitative results in Fig.~\ref{fig:sparse_view}. Our method outperforms all the baselines, especially when extrapolating from the provided two views, and can generate photorealistic novel views even if the overlap between two views is small, thanks to the strong prior from the pre-trained video generation models.

\subsection{Novel View Synthesis for Driving Simulation}
\label{sec:exp_dense_view}

\begin{figure}
  \centering
  \vspace{-2mm}
  \includegraphics[width=\linewidth]{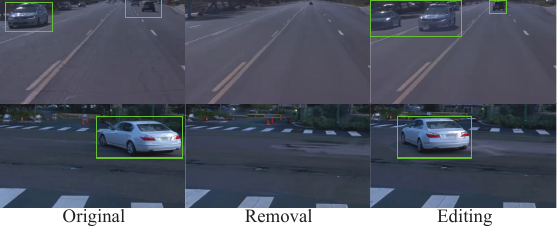}
  \vspace{-7mm}
  \caption{\footnotesize Qualitative results on 3D editing for driving scene. We remove and modify the trajectory of cars from the original scene.}
    \label{fig:editing}
    \vspace{-3mm}
\end{figure}

\begin{table}[t]
\begin{center}
\resizebox{0.9\linewidth}{!}{
\begin{tabular}{lcccc}
\toprule
\multicolumn{1}{l}{Method} & $y \pm 0.0$m & $y \pm 1.0$m & $y \pm 2.0$m & $y \pm 4.0$m \\
\midrule
Nerfacto~\cite{nerfstudio} & 48.34 & 67.77 & 80.41 & 112.40\\
3D-GS~\cite{kerbl3Dgaussians} & 34.81 & 53.85 & 61.78 & 81.26 \\
\ourModel     & \textbf{7.93} & \textbf{18.19} & \textbf{25.11} & \textbf{35.33}\\
\bottomrule
\end{tabular}
}
\end{center}
\vspace{-6mm}
\caption{\footnotesize
Quantitative results of FID~\cite{heusel2017gans} for NVS on driving scene. \ourModel significantly outperforms the baselines, especially when generating novel views that are far away from the original trajectory.}
\label{table:nerf}
\vspace{-4mm}
\end{table}

Simulating real-world driving scenes along a novel trajectory that is different from captured videos is a cornerstone for training autonomous vehicles. \ourModel can be applied.

\parahead{Evaluation and Baselines}
We compare \ourModel to two representative scene reconstruction methods, Nerfacto~\cite{nerfstudio} and 3DGS~\cite{kerbl3Dgaussians}. For a fair comparison, we filter out 18 static scenes from the validation set.
For evaluation, we create the novel trajectories by horizontally shifting from the original trajectory of the frontal camera and varying the deviation. We report  FID~\cite{heusel2017gans} for this evaluation since there is no ground truth for novel trajectories.

\parahead{Results}
As shown in Table~\ref{table:nerf}, our method achieves significantly better FID scores than reconstruction methods on driving scenarios. This is because reconstruction methods struggle to recover the scene structures from the sparsely observed views in the driving scenario. Thus, the rendering quality degrades significantly when the rendering camera moves away from the original trajectory, as shown in Fig.~\ref{fig:dense_view}.

\parahead{3D Editing}
Our explicit 3D cache naturally lends itself to 3D editing. As shown in Fig.~\ref{fig:editing}, we can remove the 3D cars, modify the trajectory of the car, and generate a plausible re-simulation video of the driving scene using \ourModel.

\subsection{Monocular Dynamic Novel View Synthesis}
\label{sec:exp_dynamic}
\begin{figure}
  \centering
  \vspace{-2mm}
  \includegraphics[width=\linewidth]{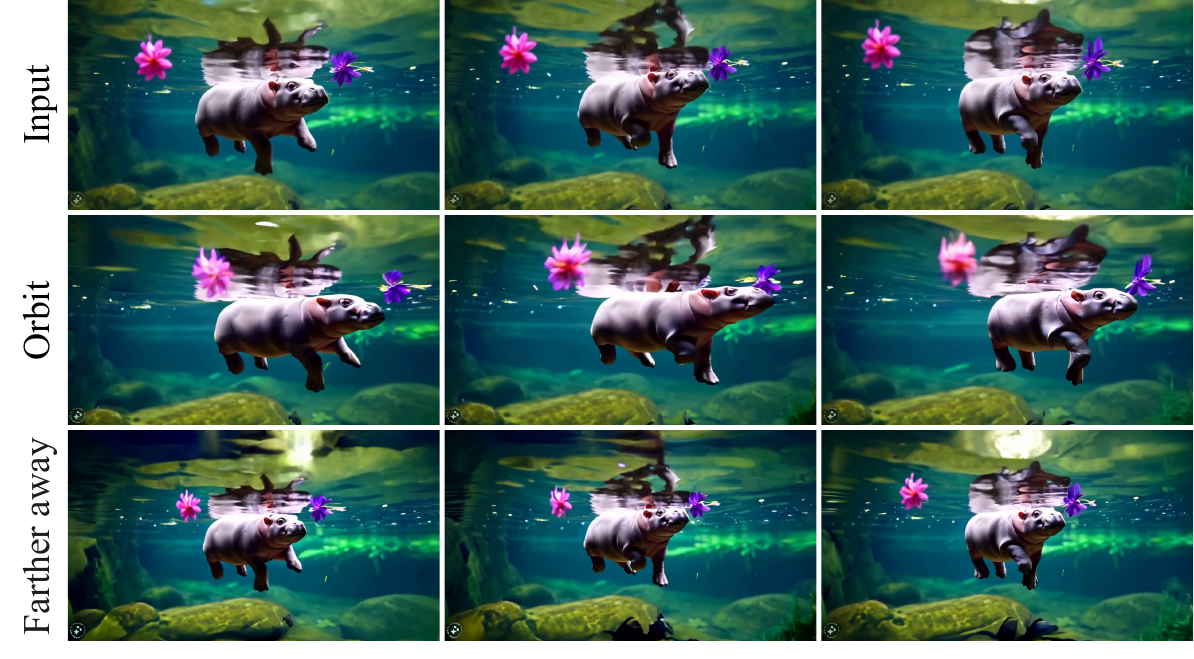}
  \vspace{-7mm}
  \caption{\footnotesize 
  Qualitative results on monocular dynamic NVS. Our model generates plausible novel camera trajectories for the given dynamic video. Note that, when zooming out from the original video, our model successfully reasons depth of field out.
}
    \label{fig:dynamic}
    \vspace{-3mm}
\end{figure}

\begin{table}[t]
\begin{center}
\resizebox{0.8\linewidth}{!}{
\begin{tabular}{lcccc}
\toprule
\multicolumn{1}{l}{Method} & PSNR $\uparrow$ & SSIM $\uparrow$ & LPIPS $\downarrow$ & FID $\downarrow$ \\
\midrule
GCD~\cite{vanhoorick2024gcd} & 19.37 & \textbf{0.67} & 0.48 & 150.64 \\
\ourModel & \textbf{19.41} & 0.63 & \textbf{0.29} & \textbf{98.58} \\
\bottomrule
\end{tabular}}
\end{center}
\vspace{-6mm}
\caption{\footnotesize 
Quantitative results on monocular dynamic NVS.}
\label{table:dynamic}
\vspace{-2mm}
\end{table}

\begin{table}[t]
\begin{center}
\resizebox{0.85\linewidth}{!}{
\begin{tabular}{cccc}
\toprule
\multicolumn{1}{l}{Noise Ratio} & PSNR $\uparrow$ & SSIM $\uparrow$ & LPIPS$\downarrow$ \\
\midrule
$0 \%$ & 24.08 / 21.56 & 0.86 / 0.83 & 0.12 / 0.15 \\
$3 \%$ & 22.39 / 21.00 & 0.83 / 0.81 & 0.16 / 0.18 \\
$10 \%$ & 20.85 / 19.64 & 0.79 / 0.76 & 0.19 / 0.22  \\
$30 \%$ &  18.52 / 17.91 & 0.72 / 0.70 & 0.29 / 0.31 \\
\bottomrule
\end{tabular}
}
\end{center}
\vspace{-5mm}
\caption{\footnotesize Quantitative results on the robustness analysis on noisy depth estimation. 
The two values in each table cell represent the interpolation and extrapolation results, respectively. 
}
\label{table:robust}
\end{table}

Given a monocular video of a dynamic scene, \ourModel is able to ``rerender'' that video along a new camera trajectory.

\parahead{Evaluation and Baselines}
We evaluate on the 20 held-out test scenes of Kubric dataset released by GCD~\cite{vanhoorick2024gcd} and compare them to GCD. We use the publicly released checkpoint trained on Kubric datasets. Since GCD is only trained at a resolution of 256x384, we upsample its predictions to the same resolution as our method for a fair comparison. 

\parahead{Results}
We provide quantitative results in Table~\ref{table:dynamic}, with qualitative results in the Supplement. Our method demonstrates superior performance in preserving object details and dynamics in input videos, and precisely aligns with the new camera motion from the users with a 3D cache.

\parahead{Out-of-Domain Results} We further qualitatively evaluate \ourModel on dynamic videos generated by Sora~\cite{sora} and MovieGen~\cite{moviegen}, and provide the results in Fig.~\ref{fig:dynamic}.  \ourModel generates photorealistic videos that preserve the 3D content and align with the new camera motion.
We refer the readers to the supplement video for full results. 

\subsection{Ablation Study}
\label{sec:ablation_study}
We ablate our method in two ways: First, different strategies to fuse the point cloud as discussed in Sec.~\ref{sec:fuse_3D_cache}, and second, the robustness to noisy depth estimation. 
We follow the experiment settings from Sec.~\ref{sec:exp_sparse_view}.

\begin{table}[t]
\centering
\resizebox{.9\linewidth}{!}{
\begin{tabular}{l|c|c|c}
\toprule
& PSNR & SSIM & LPIPS \\
\midrule
Explicit Fusion & 21.81 / 19.87 & 0.79 / 0.75 & 0.21 / 0.28 \\
Ours & \textbf{24.08} / \textbf{21.56} & \textbf{0.86} / \textbf{0.83} & \textbf{0.11} / \textbf{0.15} \\
\bottomrule
\end{tabular}}
\vspace{-2mm}
\caption{\footnotesize Ablation of different fusion strategies on RE10K dataset. The two values in each table cell represent the interpolation and extrapolation results, respectively.}
\label{tab:quantitative_ablation}
\end{table}

\begin{figure}
  \centering
  \vspace{-2mm}
  \includegraphics[width=\linewidth]{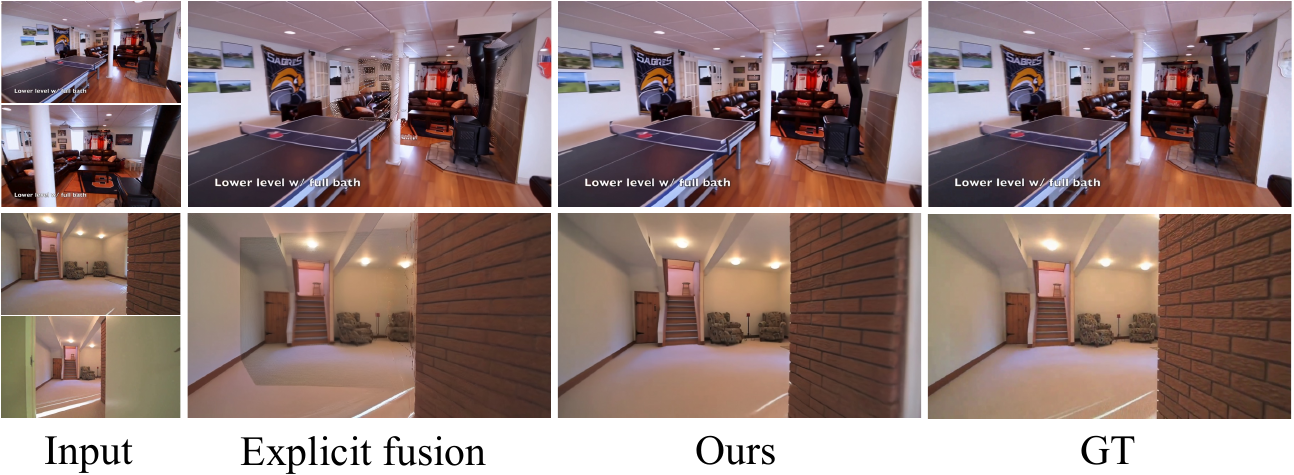}
  \vspace{-7mm}
  \caption{\footnotesize Qualitative results on ablating different fusion strategies. \ourModel can generate a realistic novel view with misaligned depth and different lighting in the input views, while the explicit fusion strategy fails.}
    \label{fig:ablation_fusion_module}
    \vspace{-4mm}
\end{figure}

\begin{figure*}
  \centering
  \vspace{-4mm}
  \includegraphics[width=\linewidth]{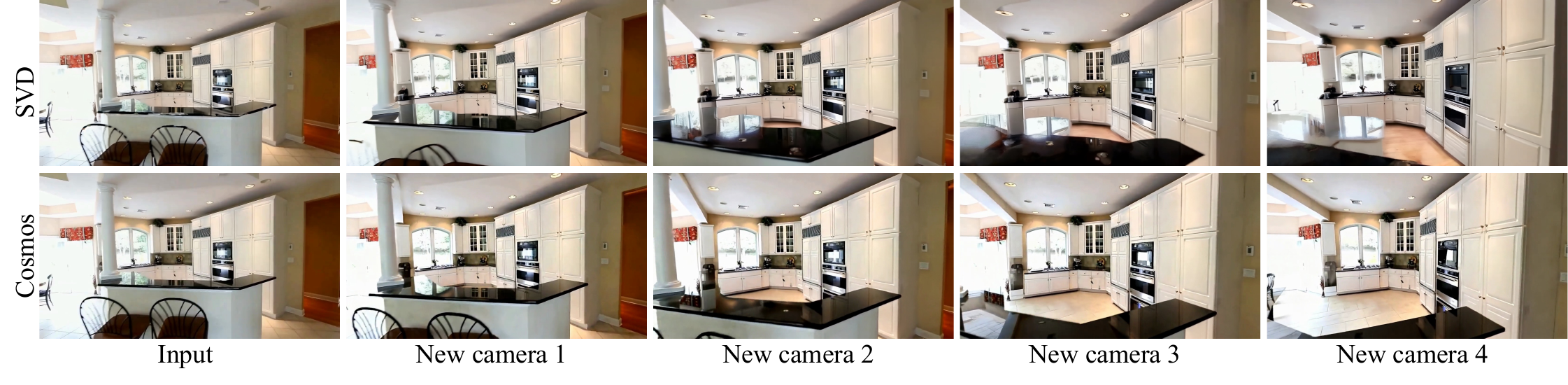}
  \vspace{-7mm}
  \caption{\footnotesize
  Qualitative comparison on using different base models: Stable Video Diffusion (SVD)~\cite{blattmann2023stable} v.s. Cosmos~\cite{cosmos}. When having a more powerful video generation model, \ourModel is able to generate more realistic output with less artifacts. Note that the slight misalignment between the two results is due to the models using different video resolutions.
  } 
    \label{fig:cosmos_svd}
    \vspace{-5.5mm}
\end{figure*}

\begin{figure}[h]
    \centering
    \begin{subfigure}[b]{0.32\linewidth}
        \includegraphics[width=\linewidth]{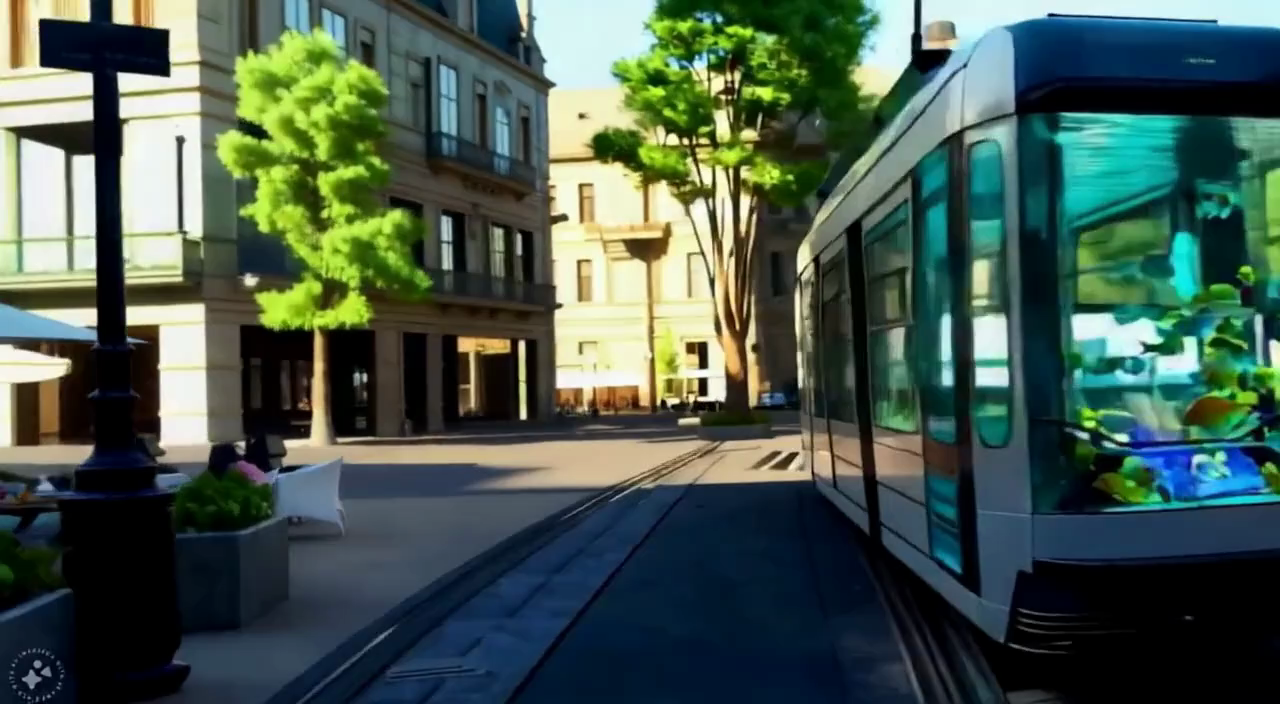}
    \end{subfigure}
    \hfill
    \begin{subfigure}[b]{0.32\linewidth}
        \includegraphics[width=\linewidth]{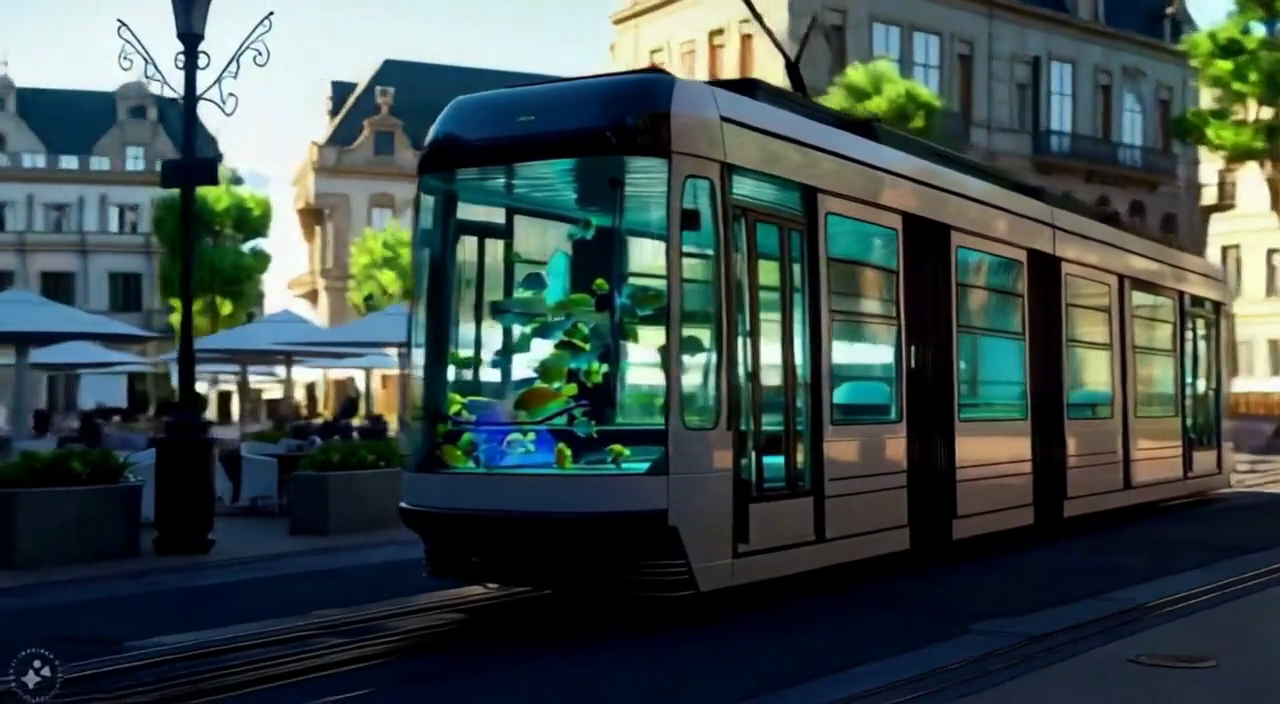}
    \end{subfigure}
    \hfill
    \begin{subfigure}[b]{0.32\linewidth}
        \includegraphics[width=\linewidth]{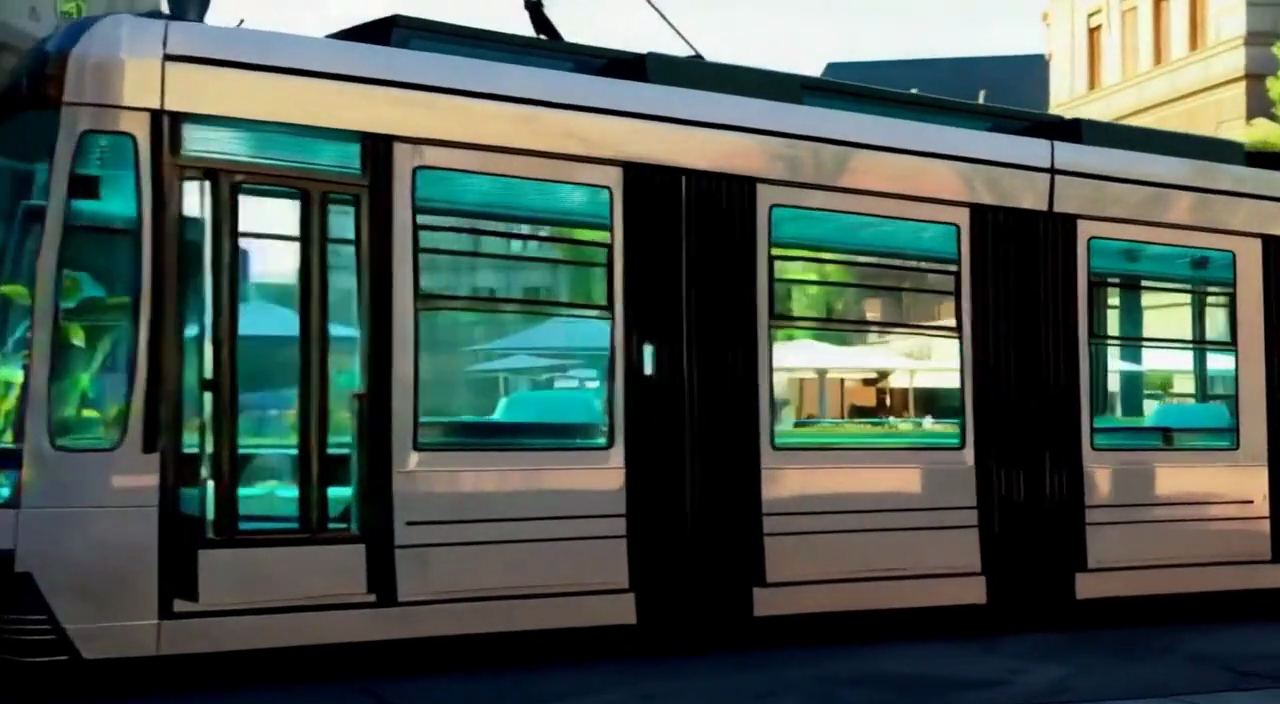}
    \end{subfigure}
    \vspace{-3mm}
    \caption{\footnotesize Example of extreme NVS using Cosmos as the base model: the input view is the middle one, and our model is capable of rotating significantly to the left and right.}
    \label{fig:extreme_NVS}
    \vspace{-4mm}
\end{figure}

\parahead{Different Fusion Strategies}
We select two input views and predict the interpolation between these two views. Our fusion strategy is compared to the explicit fusion of the point clouds from two views, in analogy to the method proposed in the concurrent work ReconX~\cite{liu2024reconx} and ViewCrafter~\cite{yu2024viewcrafter}. Qualitative examples are provided in Fig.~\ref{fig:ablation_fusion_module} with quantitative comparison in Table~\ref{tab:quantitative_ablation}. Our method can smoothly transition between two disjoint views even if the depth estimates are misaligned and the lighting is different, while explicit fusion on the point cloud suffers from severe artifacts in misalignment regions.

\parahead{Robustness to Noisy Depth Estimation}
Since our model leverages an off-the-shelf depth estimator to create the 3D cache, we need to understand the impact of error in the depth estimation. 
Given a depth estimation, we add a noise sampled from the Gaussian distribution,  $\mathcal{N}(0, s * (d_{0.95} - d_{0.05}))$. $d_{0.95}, d_{0.05}$ denote the $95$ and $5$ percentile of depth, respectively, and $s$ is the noise ratio. 
We vary the noise ratio and evaluate the capability of producing novel views. 
As shown in Table~\ref{table:robust}, when having a small amount of noise, the performance drop is negligible, and our model still performs reasonably well when the noise ratio is large (30\%).

\subsection{Extending to Advanced Video Diffusion Model}
We further replace the Stable Video Diffusion model~\cite{blattmann2023stable} with Cosmos~\cite{cosmos}, a more advanced video diffusion model which has demonstrated superior performance in video generation.
We follow the same fine-tuning protocol as above. 
Specifically, we chose \texttt{Cosmos\-1.0\-Diffusion\-7B\-Video2World}\footnote{\url{https://github.com/NVIDIA/Cosmos}} as our base model and concatenate the noisy latent with the embedding of rendered frames, which are encoded by the Cosmos tokenizer.
The model is fine-tuned on both the RE10K~\cite{zhou2018stereo} and DL3DV~\cite{ling2024dl3dv} datasets for 10,000 steps with a batch size of 64. 

We provide a qualitative comparison in Figure~\ref{fig:cosmos_svd} with additional results on our webpage\footnote{\url{https://research.nvidia.com/labs/toronto-ai/GEN3C/}}.  
Results for extreme novel view synthesis are shown in Figure~\ref{fig:extreme_NVS}.

When leveraging a more powerful video diffusion model, \ourModel is able to generate videos with much higher quality, even under extreme camera viewpoint changes.
This highlights a key strength of our method: its ability to leverage continuously evolving, pre-trained video models to achieve generalizability with minimal data required for fine-tuning.

\section{Conclusion}
In this paper, we introduced \ourModel, a consistent video generative model with precise camera control.
We achieve this goal by constructing a 3D cache from seed image(s) or previously generated videos. We then render the cache into 2D videos from a user-provided camera trajectory to strongly condition our video generation, achieving more accurate camera control than previous methods. Our results are also state-of-the-art in sparse-view novel view synthesis, even in challenging settings such as driving scenes and monocular dynamic novel view synthesis.

\parahead{Limitations}
When generating videos with dynamic content, \ourModel relies on a pre-generated video to provide the motion of the objects. Generating such a video is a challenge on its own. A promising extension is to incorporate text conditioning to prompt for motion when training video generation models.

{
    \bibliographystyle{ieeenat_fullname}
    \bibliography{main}
}

\clearpage
\appendix
\appendixpage
In the supplement, we provide additional details of our method in Sec.~\ref{suppl_sec:method} and experiments in Sec.~\ref{suppl_sec:exp}. We provide more qualitative results in Sec.~\ref{suppl_sec:results}. 

\section{Method Details}
\label{suppl_sec:method}
In this section, we provide additional details on the autoregressive generation process that updates the 3D cache.

\subsection{Auto-regressive generation}

In many applications, we need to generate a video with a sequence length that is longer than the length the original video models can support. 
To achieve this, we first divide the long video into overlapping chunks of length $\videoLength$, with a one-frame overlap between consecutive chunks, and generate the frames of each chunk sequentially in an autoregressive manner.
Specifically, for the first chunk, we follow the inference pipeline described in Sec. 4.5 of the main paper to predict an RGB video. We then update the 3D cache with the frames from the first chunk prediction, which captures a new viewpoint of the scene and provides additional information not present in the original 3D cache.

To update the 3D cache, we estimate the pixel-wise depth of the last frame in the first chunk using DAV2~\cite{depth_anything_v2}, and align this depth estimation with the 3D cache by minimizing the reprojection error. Specifically, we denote the depth estimation as $\depthEstimation$ and optimize scaling $\depthScale$ and translation $\depthTranslation$ coefficients for $\depthEstimation$. We render the point cloud from the 3D cache into a depth image at the camera viewpoint of $\depthEstimation$. We render the point cloud from the 3D cache into a depth image from the camera viewpoint of $\depthEstimation$, denoted as $\depthEstimation^{\text{tgt}}$, and, similar to the main paper, render a mask $\Mask$ indicating whether each pixel is covered by the 3D cache. The optimization objective is then defined as:
\begin{equation}
    \depthScale,\depthTranslation = \argmin_{\depthScale,\depthTranslation} \left\| \left( \depthScale \cdot \depthEstimation + \depthTranslation - \depthEstimation^{\text{tgt}} \right) \cdot \Mask \right\|_2^2.
\end{equation}
The optimized $\depthScale$ and $\depthTranslation$ are applied to normalize the depth estimation $\depthEstimation$:
\begin{equation}
    \depthEstimation' = \depthScale \cdot \depthEstimation + \depthTranslation.
\end{equation}

We unproject $\depthEstimation'$ into a 3D point cloud using the camera parameters of this frame and append it to the existing 3D cache. The updated 3D cache is subsequently used to predict the second chunk of frames. This ensures that the prediction for the second chunk is informed by the first chunk, leading to consistent generation of long videos. We iterate this process for all subsequent chunks.

\section{Experimental Details}
\label{suppl_sec:exp}
In this section, we provide additional details for our experiments in the main paper.

\begin{figure*}
  \centering
  \includegraphics[width=\linewidth]{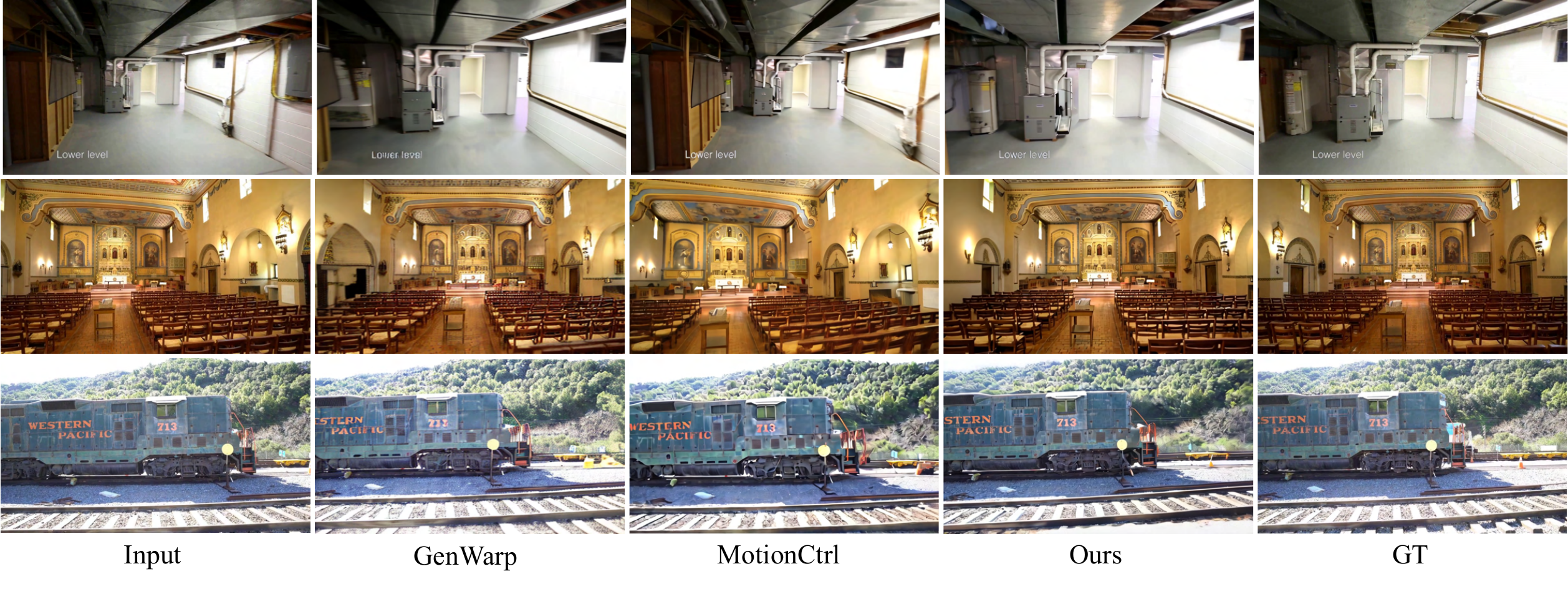}
  \vspace{-2em}
  \caption{Additional qualitative results for single-view novel view synthesis. 
  }
  \label{fig:single_view_supplement}
\end{figure*}

\begin{figure}
  \centering
  \includegraphics[width=\linewidth]{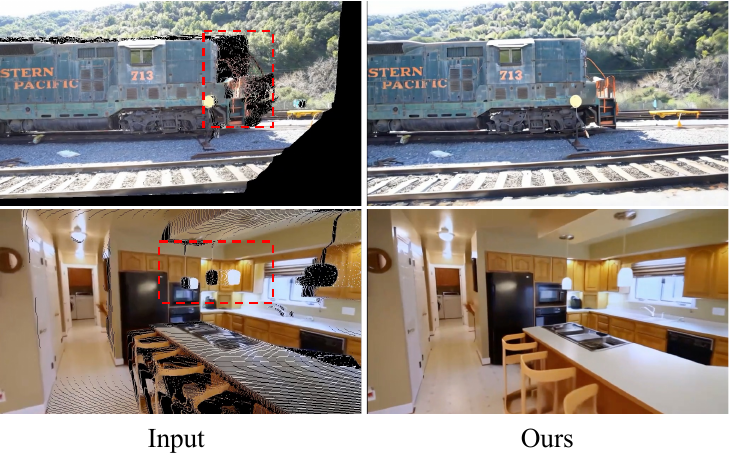}
  \vspace{-2.5em}
  \caption{Illustration of rendered depth images and model outputs. 
  Our model can fix the error in the depth projection (such as the orange handrail in the first image and the light in the second one), and generate realistic content in the missing regions (such as the inpainted railway).} 
  \label{fig:input}
\end{figure}

\begin{figure*}[t]
  \centering
  \includegraphics[width=0.9\linewidth]{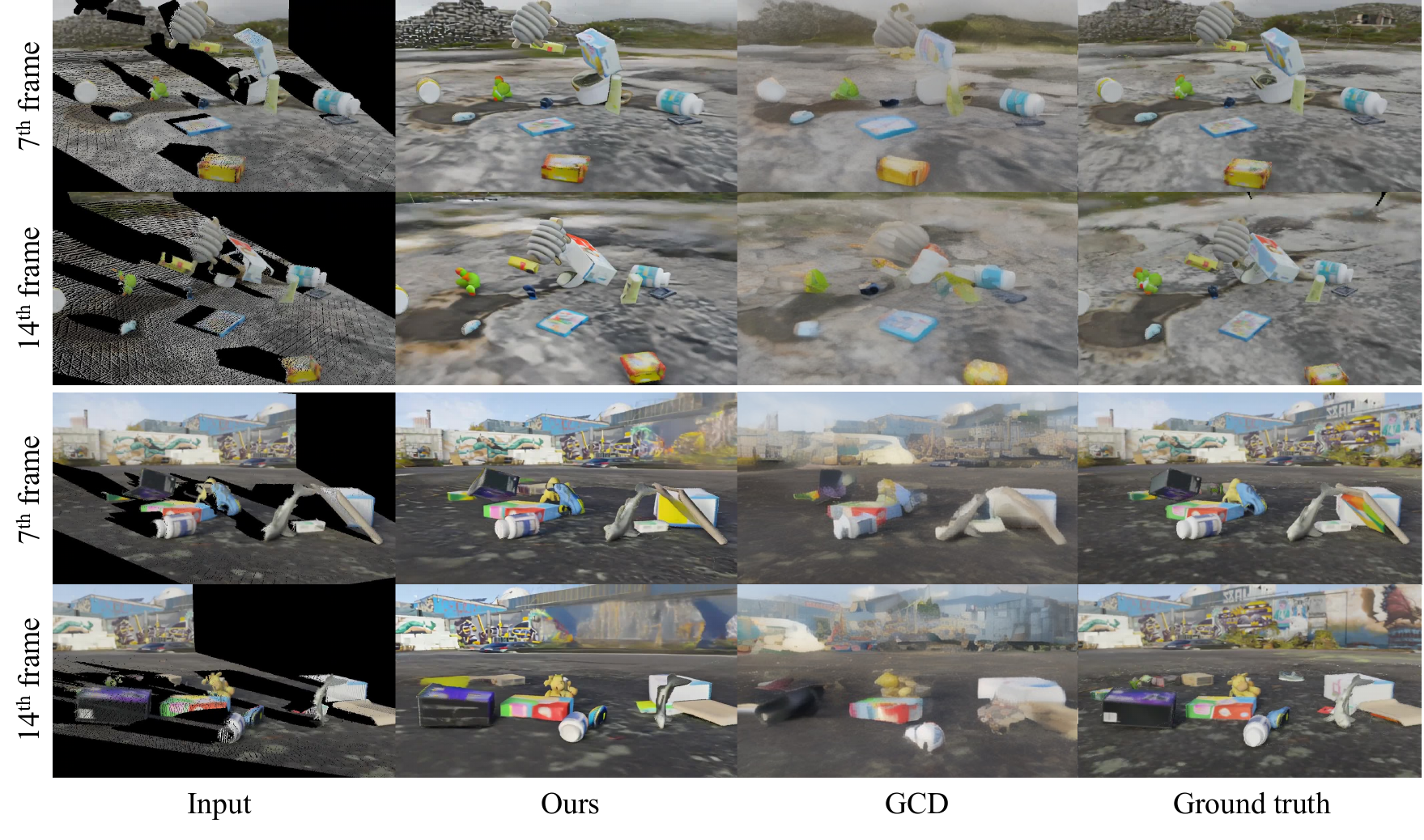}
  \caption{\footnotesize Qualitative results on Kubric4D~\cite{greff2022kubric}. Compared to GCD~\cite{vanhoorick2024gcd}, our method demonstrates superior performance in preserving object details and dynamics in input videos.}
  \label{fig:kubric}
\end{figure*}

\begin{figure}[t]
  \centering
  \includegraphics[width=\linewidth]{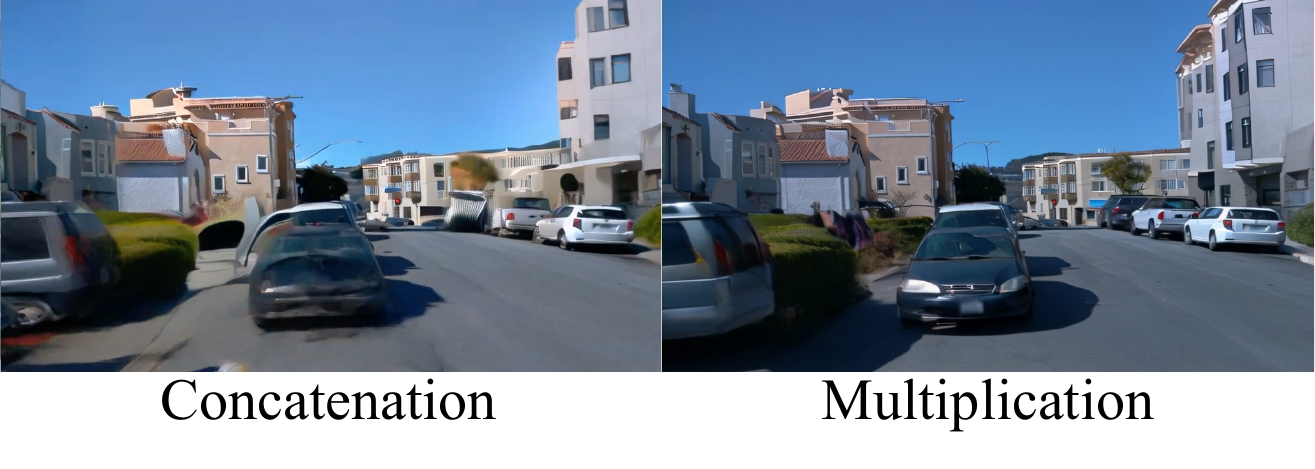}
  \vspace{-2em}
  \caption{Comparison of different strategies for incorporating masking information into the model. (Left) the mask channel is concatenated to the latent as an additional channel. (Right) the mask values are applied directly to the latent through element-wise multiplication.}
  \label{fig:mask_channel}
  \vspace{-1em}
\end{figure}

\subsection{Optimization Details}
We optimize the neural network using AdamW optimizer~\cite{loshchilov2017fixing} with the learning rate 3e-5.
During training, we apply a 15\% dropout ratio to the conditions (the rendered videos from our 3D cache and the CLIP embedding of the first frame.
We adopt a progressive training strategy for our model. Specifically, we first train the model on RE10K~\cite{zhou2018stereo} and DL3DV~\cite{ling2024dl3dv} at a resolution of $320 \times 576$ for 100K iterations. We then finetune it on all four datasets at a higher resolution of $576 \times 1024$ for another 100K iterations. In the above two stages, the sequence length is set to $14$. 
To support longer sequence lengths, we finetune the temporal layers of the video diffusion model on all four datasets for another 10K iterations at a resolution of $320 \times 576$. We first resize the video into the resolution of  $576 \times 1024$ and use center cropping to get the $320 \times 576$ video. We randomly sample a video with a sequence length ranging from $14$ to $56$ frames to finetune the temporal layer. The entire training takes around $4$ days using $32$ A100 GPUs.

\subsection{Inference Details}
We follow the practices from Stable Video Diffusion~\cite{blattmann2023stable} and use classifier-free guidance during inference with the guidance weight being 3. We run 25 diffusion steps to generate the result.

\subsection{Single-view to video generation}
We provide further details of the compared baselines in this subsection. 

\parahead{Baseline Details} 
For GenWarp~\cite{GenWarp}\footnote{\url{https://github.com/sony/genwarp}} and MotionCtrl~\cite{MotionCtrl}\footnote{\url{https://github.com/TencentARC/MotionCtrl}}, we use the official checkpoint that is trained with Stable Video Diffusion~\cite{blattmann2023stable} and evaluate on the same testing scenes as our method. Note that RE10k~\cite{zhou2018stereo} is the training dataset for two methods. 
For NVS-Solver~\cite{NVS_Solver}\footnote{\url{https://github.com/ZHU-Zhiyu/NVS_Solver}}, we use the official codebase and run the evaluation using our testing data since the model is training-free.

\subsection{Two-views NVS}
We provide further details of the compared baselines in this subsection. 

\parahead{Baseline Details} 
For PixelSplat~\cite{pixelsplat}, we take the official codebase and the released checkpoint\footnote{\url{https://github.com/dcharatan/pixelsplat}} that is trained on RE10k~\cite{zhou2018stereo} for a fair evaluation.
For MVSplat~\cite{chen2025mvsplat}, we also take the official codebase and the released checkpoint\footnote{\url{https://github.com/donydchen/mvsplat}}. 
Both baselines take images and their corresponding camera parameters as input to reconstruct 3D Gaussian Splats that can be used to render the video in the target camera trajectory. We run the two baselines on the testing data we prepared and report the results for both interpolation and extrapolation.

\subsection{NVS for driving simulation}

\parahead{Baseline Details} 
For Nerfacto, we take the official codebase released by Nerfstudio\footnote{\url{https://github.com/nerfstudio-project/nerfstudio}}, which is a state-of-the-art codebase for training Nerf. For 3DGS, we take the official codebase\footnote{\url{https://github.com/graphdeco-inria/gaussian-splatting}}.
We use all frames from three cameras in the training data to train Nerfacto and 3DGS.

\parahead{Inference Pipeline} 
With a driving video, we estimate the depth of each frame and align it with the depth scale from the Lidar point cloud. We then unproject the depth estimation for each frame. In this case, we treat each frame as a different time capture of the same scene and concatenate them along the temporal dimension of the 3D cache, since there could exist dynamic objects in the scene. 
With the new camera trajectory provided by the user, we render the 3D cache along this trajectory. 
The rendering of the 3D cache is then used to generate the video at the novel camera trajectory.
 
\subsection{Monocular Dynamic NVS}

\parahead{Inference Pipeline} 
With a monocular video of a dynamic scene, we aim to generate a video of the same scene from a different camera trajectory.
Similar to the driving simulation, we predict the depth for each frame separately and concatenate the unprojected point cloud from depth along the temporal dimension of the 3D cache. 
With the new camera trajectory provided by the user, we render the 3D cache along this new camera trajectory. 
The rendered video is fed into \ourModel to generate a dynamic video output.

\section{Additional Results}
\label{suppl_sec:results}

\subsection{Generalization with mask channel}

In Sec. 4.3 of the main paper, we discuss different strategies for incorporating mask information into the model. Here, we provide an additional qualitative comparison in Fig.~\ref{fig:mask_channel}. Concatenating the mask channel to the latent introduces additional model parameters, which do not generalize well to out-of-distribution masks during training. This issue is particularly severe in driving simulations, where ground truth views for novel trajectories (e.g., horizontal shifts from the original trajectory) are unavailable. In contrast, directly multiplying the mask with the latent demonstrates better generalization, significantly reducing artifacts when synthesizing extreme novel views.

\subsection{Single-view to video generation}

In addition to the comparison in the main paper, we provide the quantitative comparisons with GenWarp~\cite{GenWarp} and MotionCtrl~\cite{MotionCtrl} in \cref{fig:single_view_supplement}. 
We also provide the rendered depth images and the model outputs in \cref{fig:input} to demonstrate the capability of our model on both filling missing regions in the 3D cache and fixing artifacts from the rendering of the imperfect 3D cache.

\subsection{Monocular Dynamic Novel View Synthesis}

We provide the qualitative comparison with GCD~\cite{vanhoorick2024gcd} on the Kubric dataset in Fig.~\ref{fig:kubric}. Our method generates sharper video details with more object details and dynamics compared to GCD~\cite{vanhoorick2024gcd}.

\end{document}